\documentclass[a4paper,11pt,oneside]{article}
\usepackage[utf8]{inputenc}

\usepackage[english]{babel} 
\usepackage{geometry}
\usepackage{amsmath,amssymb,amsthm}
\usepackage{epsfig}
\usepackage[utf8]{inputenc}
\usepackage{textcomp}
\usepackage{algorithm}
\usepackage{color}
\usepackage{graphicx}
\usepackage{caption} 
\usepackage{subcaption}
\usepackage{bm}
\usepackage{float}
\usepackage{url}
\usepackage{dsfont}

\usepackage[noend]{algpseudocode}

\newtheorem{thm}{Theorem}[section]

\newtheorem{remark}[thm]{Remark}

\DeclareMathOperator*{\argmin}{arg\,min}

\newcommand{\Prob}{\mathbb{P}}

\newcommand{\E}{\mathbb{E}}

\newcommand{\g}[1]
{
\mathbf{ #1}
}

\newcommand{\bs}[1]
{
\boldsymbol{ #1}
}

\algnewcommand\INPUT{\item[\textbf{Input:}]}%
\algnewcommand\OUTPUT{\item[\textbf{Output:}]}%

\algdef{SE}[SUBALG]{Indent}{EndIndent}{}{\algorithmicend\ }%
\algtext*{Indent}
\algtext*{EndIndent}

\makeatletter
\def\blfootnote{\xdef\@thefnmark{}\@footnotetext}
\makeatother

\newlength\figureheight
\newlength\figurewidtht

\title{Uncertainty quantification for industrial design using dictionaries of reduced order models} 

\author{Thomas Daniel${}^{*,\dagger}$, Fabien Casenave${}^{*}$, Nissrine Akkari${}^{*}$, David Ryckelynck${}^{\dagger}$, Christian Rey${}^{*}$}

\begin{document}

\maketitle
\blfootnote{${}^{*}$ SafranTech, Rue des Jeunes Bois, Ch\^ateaufort, 78114 Magny-les-Hameaux (France).}
\blfootnote{${}^{\dagger}$ MINES ParisTech, PSL University, Centre des mat\'eriaux (CMAT), CNRS UMR 7633, BP 87, 91003 Evry (France)}

\begin{abstract}
We consider the dictionary-based ROM-net (Reduced Order Model) framework \emph{[T. Daniel, F. Casenave, N. Akkari, D. Ryckelynck, Model order reduction assisted by deep neural networks (ROM-net), Advanced modeling and Simulation in Engineering Sciences 7 (16), 2020]} and summarize the underlying methodologies and their recent improvements. The main contribution of this work is the application of the complete workflow to a real-life industrial model of an elastoviscoplastic high-pressure turbine blade subjected to thermal, centrifugal and pressure loadings, for the quantification of the uncertainty on dual quantities (such as the accumulated plastic strain and the stress tensor), generated by the uncertainty on the temperature loading field. The dictionary-based ROM-net computes predictions of dual quantities of interest for 1008 Monte Carlo draws of the temperature loading field in 2~hours and 48~minutes, which corresponds to a speedup greater than 600 with respect to a reference parallel solver using domain decomposition, with a relative error in the order of 2\%. Another contribution of this work consists in the derivation of a meta-model to reconstruct the dual quantities of interest over the complete mesh from their values on the reduced integration points.
\end{abstract}


\section{Introduction}

Numerical simulation is vastly used in the industry when designing a new mechanical part. Uncertainty quantification is used to compute the influence of a poorly known or uncontrolled parameter, like dispersion within manufacturing tolerances.
Procedures that rely on the Monte Carlo method require solving the corresponding problem for many values of the parameter.
With modeling and simulation progresses, meshes are getting larger and models more complex, leading to increased duration for the corresponding computations.
For these reasons, many methods have been proposed to replace these reference models with fastly computed approximations.

In this work, we consider an industrial model of an elastoviscoplastic high-pressure turbine blade subjected to thermal, centrifugal and pressure loadings, for the quantification of the uncertainty of dual quantities (such as the accumulated plastic strain and the stress tensor), generated by the uncertainty on the temperature loading. 
Computing the fatigue lifetime of one such blade requires simulating its behavior until the stabilization of the mechanical response, which last several weeks using Abaqus~\cite{abaqus} because of the size of the mesh, the complexity of the constitutive equations, and the number of loading cycles in the transient regime. With such a computation time, uncertainty quantification with the Monte Carlo method is unaffordable. In addition, such simulations are too time-consuming to be integrated in design iterations, which limits them to the final validation steps, while the design process still relies on simplified models. Accelerating these complex simulations is a key challenge while maintaining a satisfying accuracy, as it would provide useful numerical tools to improve design processes and quantify the effect of the uncertainties on the environment of the system.


Simulations are accelerated using a dictionary of reduced order models, with a classifier able to select which local reduced order model use for a new temperature loading. The framework is called ROM-net~\cite{ROM-net}. A dataset of 200 solutions is computed in a Finite Element approximation space of dimension in the order of the million, for various instances of the temperature field loading in parallel in 7~days and 9~hours on 48~cores.
These solutions are computed over 11 time steps in the first cycle, using a scalable Adaptive MultiPreconditioned FETI (AMPFETI) solver~\cite{feti} in Z-set finite-element software~\cite{zset}. The dataset is partitioned into two clusters using a k-medoids algorithm with a Reduced Order Model (ROM)-oriented dissimilarity measure in 5~minutes; the corresponding local ROMs, using Proper Orthogonal Decomposition (POD)~\cite{cordier:hal-00417819, RowleyPODGalerkin} and Empirical Cubature Method (ECM)~\cite{ECM}, are trained in 2~hours and 30~minutes. An automatic reduced model recommendation procedure, allowing to decide which local ROM to use for a new temperature loading, is trained in the form of a logistic regression classifier in 16 minutes. A contribution of this work is the use of a meta-model to reconstruct the dual quantities of interest over the complete mesh from their values on the reduced integration points, in the form of a multi-task Lasso, which takes 1~hour to train for 14~dual fields. The uncertainties on dual quantities of interest, such as the accumulated plastic strain and the stress tensor, are quantified by using our trained ROM-net on 1008 Monte Carlo draws of the temperature loading field in 2~hours and 48~minutes, which corresponds to a speedup greater than 600 with respect to our highly optimized domain decomposition AMPFETI solver. Expected values for the Von Mises stress and the accumulated plastic strain have 0.99-confidence intervals' width of respectively 1.66\% and 2.84\% (relative to the corresponding prediction for the expected value). As a validation stage, 20 reference solutions are computed on new temperature loadings, and dual quantities of interest are predicted with relative accuracy in the order of 1\% to 2\%, while the location of the maximum value is perfectly predicted.

In Section~\ref{sec:methodo}, a summary of the results and the methodology constructed in our team's previous work on dictionary-based ROM-net is provided. Details are given on the reduced order modeling strategy and the concept of ROM-net is recalled. References on alternative methods from the literature are also given. Section~\ref{sec:indusCase} contains a description of the industrial dataset, the hypotheses of the model, and the objective of the present study. The proposed workflow for uncertainty quantification is then applied on this industrial configuration in Section~\ref{sec:indusStudy}. Finally, conclusions are drawn in Section~\ref{sec:conclusion}.

\section{Description of the methodology}
\label{sec:methodo}

This section provides elements from the literature and from our previous works to deal with uncertainty quantification in an industrial context using physical reduced order modeling.

\subsection{Reduced order modeling of nonlinear parametrized partial differential equations}
\label{sec:nonlinearROM}

Consider a nonlinear physical problem described by the following parametrized differential equation:\begin{equation}
\mathcal{D}(u;x) = 0,
\label{PDE}
\end{equation} where $u$ is the primal variable belonging to a Hilbert space $\mathcal{H}$, $x$ denotes the parameters of the problem, and $\mathcal{D}$ is an operator involving a differential operator and operators for initial conditions and/or boundary conditions. Here, the time is considered as a parameter and is included in the definition of $x$.
The \textit{solution manifold} $\mathcal{M}$ is defined by $\mathcal{M} := u(\mathcal{X}) = \{ u(x) \ | \ x \in \mathcal{X} \}$.
Classically, the problem is written on a geometrical support discretized by a mesh, and the solution is sought in a finite dimensional space, e.g. the finite element space $Span\left\{ \phi_{i} \right\}_{1 \leq i \leq \mathcal{N}}$: this defines the High Fidelity Model (HFM). Here, in the target application, $\mathcal{N}$ is in the order of the million.

Model order reduction~\cite{10.5555/2568435, keiper2018reduced} is a discipline in numerical analysis consisting in replacing a computationally expensive high fidelity model by a fast ROM to calculate approximate solutions of the considered physical problem~\eqref{PDE}. A ROM can be either a data-driven metamodel (or surrogate model) calibrated with a regression algorithm, or a physics-based model obtained by numerical methods such as the Proper Generalized Decomposition~\cite{PGDreview, PGDbook}, the Reduced Basis Method (RBM)~\cite{RBmethodPrudhomme, RozzaReducedBasis}, and the POD-Galerkin method~\cite{cordier:hal-00417819, RowleyPODGalerkin}, among others. 
It is generally used for parametrized equations whose solution must be known for different points in the parameter space. As in machine learning, a model order reduction procedure starts by a \textit{training phase} (or \textit{offline stage}) where the ROM is built from some training data. The ROM is then used on test data in an \textit{exploitation phase} (or \textit{online stage}). In the training phase, high-fidelity solutions, called \textit{snapshots}, are computed with the HFM for different points of the parameter space to get a sampled representation of the solution manifold. The model order reduction algorithm analyzes these snapshots to learn how the solution is affected by parameter variations. Given the cost of computing snapshots in the training phase, a ROM is profitable only if it is extensively used in the exploitation phase.

RBM and POD-Galerkin are also called \textit{projection-based} since they consist in applying the Galerkin method on a reduced order basis, which can be costly for certain parameter dependencies and nonlinear problems. In such cases, a second reduction stage is necessary.
\textit{Hyper-reduction} was initially the name of a method proposed in~\cite{Ryckelynck2005} in 2005, but this term has been extended to refer to all the methods proposing a second reduction stage. Hyper-reduction methods include the Empirical Interpolation Method (EIM, \cite{EIM}), the Missing Point Estimation (MPE, \cite{MPE}), the A~Priori Hyper-Reduction (APHR, \cite{Ryckelynck2005}), the Best Point Interpolation Method (BPIM, \cite{BPIM}), the Discrete Empirical Interpolation Method (DEIM, \cite{DEIM}), the Gauss-Newton with Approximated Tensors (GNAT, \cite{GNAT}), the Energy-Conserving Sampling and Weighting (ECSW, \cite{ECSW}), the Empirical Cubature Method (ECM, \cite{ECM}), and the Linear Program Empirical Quadrature Procedure LPEQP, \cite{LPempirical}). Hyper-reduction techniques implicitly assume that the physics model is based on \textit{local} constitutive laws. A constitutive model is \textit{local} if its equations evaluated at a given point $\bs{\xi}\in\Omega$ only involve variables evaluated at $\bs{\xi}$.

We use the reduced order modeling framework developed in our previous work~\cite{CasenaveIJNME} for elastoviscoplastic structural mechanics, whose training phase is decomposed in three steps:
\begin{itemize}
\item \textbf{Data generation:} snapshots $u(x_n)$, $1\leq n\leq N_s$, $N_s$ being the number of available snapshots, are computed with the high-fidelity model and provide information about how the physical system reacts to changes of the parameter $x$. In our case, the finite-element solver \textit{Z-set}~\cite{zset} is used.
\item \textbf{Data compression:} a Reduced Order Basis (ROB) is constructed by looking for a hidden low-rank structure in the snapshots. We apply the snapshot-POD, which consists in (i) computing the snapshot correlation matrix $\displaystyle C_{n,m}=\int_{\Omega}u(x_n)\cdot u(x_m)$, $1\leq n,m\leq N_s$, (ii) retaining the eigenvalue/eigenvector pairs of $C$ associated to the highest eigenvalues: $(\xi_i,\lambda_i)$, $1\leq i\leq N$, and (iii) recombining them with the snapshots to create the ROB: $\displaystyle\psi_i(x)=\frac{1}{\sqrt{\lambda_i}}\sum_{n=1}^{N_n}u(x_n){\xi_i}_n $, $1\leq i\leq N$. When the quantities of interest are dual variables, which is the case in the industrial application considered here, ROBs are also constructed for the corresponding fields.
\item \textbf{Operator compression:} this step contains the operations that guarantee the efficiency of the reduced order model in the exploitation phase. For nonlinear problems, applying the Galerkin method in the online phase on the ROB requires to compute integrals over the mesh. In our case, we use the ECM~\cite{ECM} to replace these costly integrals by reduced quadrature schemes trained over the snapshots at our disposal, hence tailored for our considered problem.
\end{itemize}

Then, the online stage consists in assembling a variational formulation of Equation~\eqref{PDE} on the ROB, using a reduced quadrature scheme, and solving it using a Newton algorithm. Since the constitutive laws are only computed at the locations of the integration points of the reduced quadrature scheme, the dual quantities have to be reconstructed over the complete mesh. This can be done using the Gappy-POD and the ROBs of the corresponding dual fields. See~\cite{CasenaveIJNME} for more details on the reduced order modeling framework we used. In Section~\ref{sec:GappyMetaModel}, we propose to replace this last reconstruction stage by a meta-model.

\subsection{Dictionary of reduced order models}

This section introduces the concept of ROM-net as a dictionary of ROMs, and provides elements on its implementation.

\subsubsection{Nonreducible problems}

A ROM is an approximation of our considered HFM.
We define the speedup of the ROM as the ratio of the computation time of the HFM to the computation time of the ROM. Consider a set of snapshots generated using the high-fidelity model over a sampling of the parameter domain. The parametrized problem is said nonreducible when applying a linear data compression over this set of snapshots leads to a ROB containing too many vectors for the online problem to feature an interesting speedup.
Formally, this happens when the Kolmogorov N-width $d_{N}(\mathcal{M})$ decreases too slowly with respect to $N$, where $N$ is the cardinality of the ROB,
\begin{equation}
d_{N}(\mathcal{M}) := \underset{\mathcal{H}_{N} \in \textrm{Gr}(N,\mathcal{H})}{\inf} \  \underset{u \in \mathcal{M}}{\sup}  \ \underset{v \in \mathcal{H}_N}{\inf} || u - v ||_{\mathcal{H}},
\label{KolmogorovWidth}
\end{equation} 
with the Grassmannian $\textrm{Gr}(N,\mathcal{H})$ being the set of all $N$-dimensional subspaces of $\mathcal{H}$ and $\mathcal{H}_{N} \in \textrm{Gr}(N,\mathcal{H})$ the subspace spanned by the considered ROB.
Qualitatively, the solution manifold $\mathcal{M}$ covers too many independent directions to be embedded in a low-dimensional subspace. To address this issue, several techniques have been developed:
\begin{itemize}
\item Problem-specific methods tackle the difficulties of some specific physics problems that are known to be nonreducible, such as advection-dominated problems which have been largely investigated, for instance in~\cite{PhysRevE.89.022923, doi:10.1137/17M1140571, MadayConvection}.
\item Online-adaptive model reduction methods update the ROM in the exploitation phase by collecting new information online as explained in~\cite{Zimmermann2017}, in order to limit extrapolation errors when solving the parametrized governing equations in a region of the parameter space that was not explored in the training phase. The ROM can be updated for example by querying the high-fidelity model when necessary for basis enrichment~\cite{Ryckelynck2005, 10.1145/1618452.1618469, doi:10.1137/151003660, CasenaveAkkari, he2020insitu}.
\item ROM interpolation methods~\cite{Interpolation0, Interpolation1, Interpolation2, Interpolation3, Interpolation4, Interpolation5, Interpolation6, doi:10.2514/1.J050233, AMSALLEM2016373, Interpolation7, Interpolation8, CHOI2020109787} use interpolation techniques on Grassmann manifolds or matrix manifolds to adapt the ROM to the parameters considered in the exploitation phase by interpolating between two precomputed ROMs.
\item Dictionaries of basis vector candidates enable building a parameter-adapted ROM in the exploitation phase by selecting a few basis vectors. This technique is presented in~\cite{doi:10.1137/120873868, Kaulmann2012ONLINEGR} for the Reduced Basis method.
\item Dictionaries of ROMs rely on the construction of several local ROMs adapted to different regions of the solution manifold. These local ROMs can be obtained by partitioning the time interval~\cite{Drohmann2010, Dihlmann2011}, the parameter space~\cite{Drohmann2010, Eftang2010, Haasdonk2011, LDEIM2014, he2020insitu, doi:10.2514/6.2020-0418, kapteyn2020physicsbased}, or the solution space~\cite{LocalROB, localROB2, LDEIM2014, Amsallem2015localHROM, RyckelynckComputerVision, ROM-net, Grimberg2020, CRMECA_2020__348_10-11_911_0}.
\item Nonlinear manifold ROM methods~\cite{LEE2020108973, lee2020DeepConservation, KimChoi2020} learn a nonlinear embedding and project the governing equations onto the corresponding approximation manifold, by means of a nonlinear function mapping a low-dimensional latent space to the solution space.
\end{itemize}

Our framework focuses on dictionaries of ROMs, where the solution manifold is partitioned to get a collection of subsets $\mathcal{M}_k \subset \mathcal{M}$ that can be covered by a dictionary of low-dimensional subspaces, enabling the use of local linear ROMs. If $\{ \mathcal{M}_k \}_{k \in [\![ 1;K ]\!]}$ is a partition of $\mathcal{M}$, then:\begin{equation}
\forall k \in [\![ 1;K ]\!], \ \forall N \in \mathbb{N}^{*}, \quad d_{N}(\mathcal{M}_k) \leq d_{N}(\mathcal{M}).
\end{equation}
For a given number $K$ of subsets, two partitions can be compared on the basis of the ratios $d_{N}(\mathcal{M}_k) / d_{N}(\mathcal{M})$.

\subsubsection{Dictionary-based ROM-nets}
\label{sec:dicROMnets}

\begin{figure}[!h]
\centering
\includegraphics[scale=0.31]{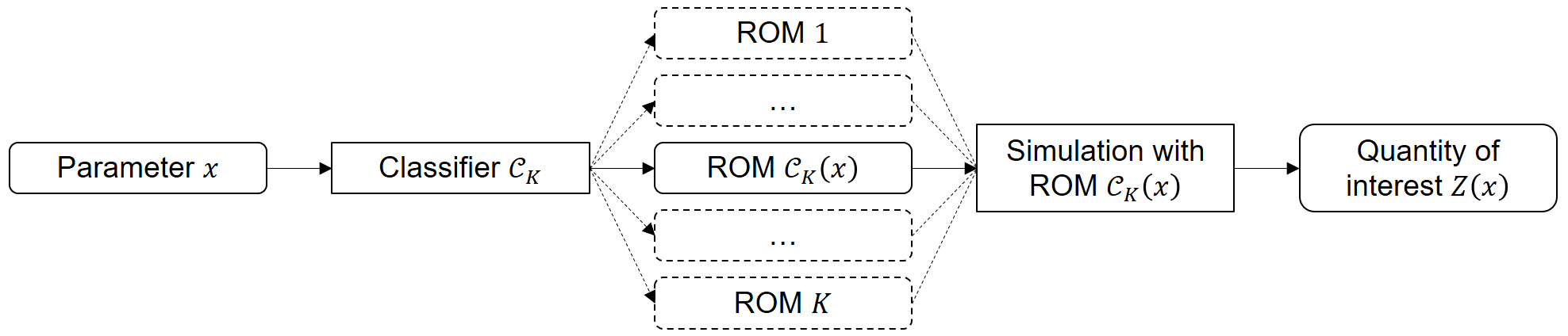}
\caption{Exploitation phase of a dictionary-based ROM-net. $K$ local ROMs are combined with a classifier $\mathcal{C}_K$ for automatic ROM $\mathcal{C}_K(x)$ recommendation, used to predict the quantity of interest $Z(x)$.}
\label{ROM-net_online}
\end{figure}

We introduced the concepts of ROM-net and dictionary-based ROM-net in~\cite{ROM-net}, where rigorous definitions can be found. 
Suppose we dispose of an already computed dictionary of ROMs for the parametrized problem~\eqref{PDE}, where each element of the dictionary is a ROM that can approximate the problem on a subset of the solution manifold~$\mathcal{M}$.
A dictionary-based ROM-net is a machine learning algorithm trained to assign the parameter $x\in\mathcal{X}$ to the ROM of the dictionary leading to the most accurate reduced prediction. This assignment, called model recommendation in~\cite{RyckelynckComputerVision}, is a classification task, see Figure~\ref{ROM-net_online}. 

The dictionary of ROMs is constructed in a clustering stage, during which snapshots are regrouped depending on their respective proximity on $\mathcal{M}$, in the sense of a particular dissimilarity measure we introduced in~\cite{daniel2021physicsinformed}.
The dissimilarity between two parameter values $x, x'\in\mathcal{X}$, denoted by ${\delta}(x, x')$, involves the sine of the principal angles between subspaces associated to the solutions of the HFM $u(x), u(x')\in\mathcal{M}$, see~\cite[Definition~3.11]{daniel2021physicsinformed}.
For this reason, the clustering is coined \textit{physics-informed}.
We refer to the remaining of~\cite{daniel2021physicsinformed} for the description of a practical efficiency criterion of the dictionary-based ROM-net, which enables to decide, before the computationally costly steps of the workflow, if a dictionary of ROMs is preferable to one global ROM, and how to calibrate the various hyperparameters of the ROM-net. 



\begin{remark}[Importance of the classification]
We use a representative-based clustering algorithm, namely k-medoids. One could argue that the classification step can be replaced by choosing the cluster $k$ for which the dissimilarity measure ${\delta}(x, \tilde{x}_k)$ between the parameter $x$ and the cluster medoid $\tilde{x}_k$ is the smallest. However, we recall that the computation of the dissimilarity measure requires solving the HFM at the parameter value $x$, which would render the complete model reduction framework useless. Hence, the classification step enables to bypass this HFM solve and directly recommend the appropriate local ROM.
\end{remark}

The training of the classifier can be difficult when working with physical fields: simulations are costly, data are in high dimension and classical data augmentation techniques for images cannot be applied. Hence, we can consider replacing the HFM by an intermediate-fidelity solver for generating the data needed for the training of the classifier, by considering coarser meshes and fewer time steps. We precise that the HFM should be used at the end for generating the data required in the training of the local ROMs. We propose in~\cite{mca26010017} improvements for the training of the classifier in our context by developing a fast variant of the mRMR~\cite{mRMR2005} feature selection algorithm, and new class-conserving transformations of our data, acting like a data augmentation procedure. In this work, we use the same model for generating the data used in the training of the clustering and classifier and for constructing the local reduced-order models: there is no intermediate-fidelity solver.

\subsection{Uncertainty quantification}

The parameter is modeled by a random variable. The uncertainty quantification consists in a Monte Carlo procedure where values of the parameter are drawn from the distribution of this random variable, and using the trained dictionary-based ROM-net to select a local ROM and predicting the corresponding quantity of interest. Statistical estimators of quantities depending on the solution of our physical problem can then be efficiently computed.
Our objective is to apply the presented methodology to a real industrial case: quantifying the uncertainties on dual quantities of interest generated by the uncertainty of the temperature loading, in a high-pressure (HP) turbine blade elastoviscoplastic cyclic mechanical computation.

\section{Industrial context}
\label{sec:indusCase}

This section presents the industrial test case of interest. It consists in predicting the mechanical behavior of a HP turbine blade in an aircraft engine with uncertainties on the thermal loading. The industrial context and the models for the mechanical behavior and the thermal loading are presented, with a particular emphasis on the assumptions that have been made.

For confidentiality reasons, mesh sizes and numerical values corresponding to the industrial dataset are not given. Reproducible data are available on request for the numerical example proposed in~\cite{ROM-net}. The accuracy of the predictions made by our methodology are given in the form of relative errors.

\subsection{Thermomechanical fatigue of high-pressure turbine blades}

High-pressure turbine blades are critical parts in an aircraft engine. Located downstream of the combustion chamber, they are subjected to extreme thermomechanical loadings resulting from the combination of centrifugal forces, pressure loads, and hot turbulent fluid flows whose temperatures are higher than the material's melting point. The repeating thermomechanical loading over time progressively damages the blades and leads to crack initiation under thermomechanical fatigue. Predicting the fatigue lifetime is crucial not only for safety reasons, but also for ecological issues, since reducing fuel consumption and improving the engine's efficiency requires increasing the temperature of the gases leaving the combustion chamber.

High-pressure turbine blades are made of monocrystalline nickel-based superalloys that have good mechanical properties at high temperatures. To reduce the temperature inside this material, the blades contain cooling channels where flows relatively fresh air coming from the compressor. In addition, the blade's outer surface is protected by a thin thermal barrier coating. In spite of these advanced cooling technologies, the rotor blades undergo centrifugal forces at high temperatures, causing inelastic strains. Under this cyclic thermomechanical loading repeated over the flights, the structure has a viscoplastic behavior and reaches a viscoplastic stabilized response, where the dissipated energy per cycle still has a nonzero value. This is called \textit{plastic shakedown}, and leads to \textit{low-cycle fatigue}. At cruise flight, the persistent centrifugal force applied at high temperature induces progressive (or time-dependent) inelastic deformations: this phenomenon is called \textit{creep}. In addition, the difference between gas pressures on the extrados and the intrados of the blade generates bending effects. Environmental factors may also locally modify the chemical composition of the material, leading to its \textit{oxidation}. As oxidized parts are more brittle, they facilitate crack initiation and growth. \textit{Thermal fatigue} resulting from temperature gradients is another life-limiting factor. Temperature gradients make colder parts of the structure prevent the thermal expansion of hotter parts, creating thermal stresses. Due to their higher temperatures, the hot parts are more viscous and have a lower yield stress, which make them prone to develop inelastic strains in compression. When the temperature cools down after landing, tensile \textit{residual stresses} appear in parts which were compressed at high temperatures and favor crack nucleation. Given the complex temperature field resulting from the internal cooling channels and the turbulent gas flow, thermal fatigue has a strong influence on the turbine blade's lifetime. In particular, during transient regimes such as take-off, an important temperature gradient appears between the leading edge and the trailing edge of the blade, since the latter has a low thermal inertia due to its small thickness and thus warms up faster.

In short, the behavior of a high pressure turbine blade results from a complex interaction between low-cycle fatigue, thermal fatigue, creep, and oxidation. Due to the cost and the complexity of experiments on parts of an aircraft engine, numerical simulations play a major role in the design of high-pressure turbine blades and their fatigue lifetime assessment. All this knowledge have been learned by scientist and engineers during last decades. In the proposed approach to machine learning for model order reduction, all this knowledge is preserved in local ROMs. It is even more than that, the uncertainty propagation comes to complete this valuable traditional knowledge. We do not expect from artificial intelligence to learn everything in our modeling process.

\subsection{Industrial dataset and objectives}

\subsubsection{Industrial problem}

\begin{figure}[!h]
\centering
\includegraphics[scale=0.4]{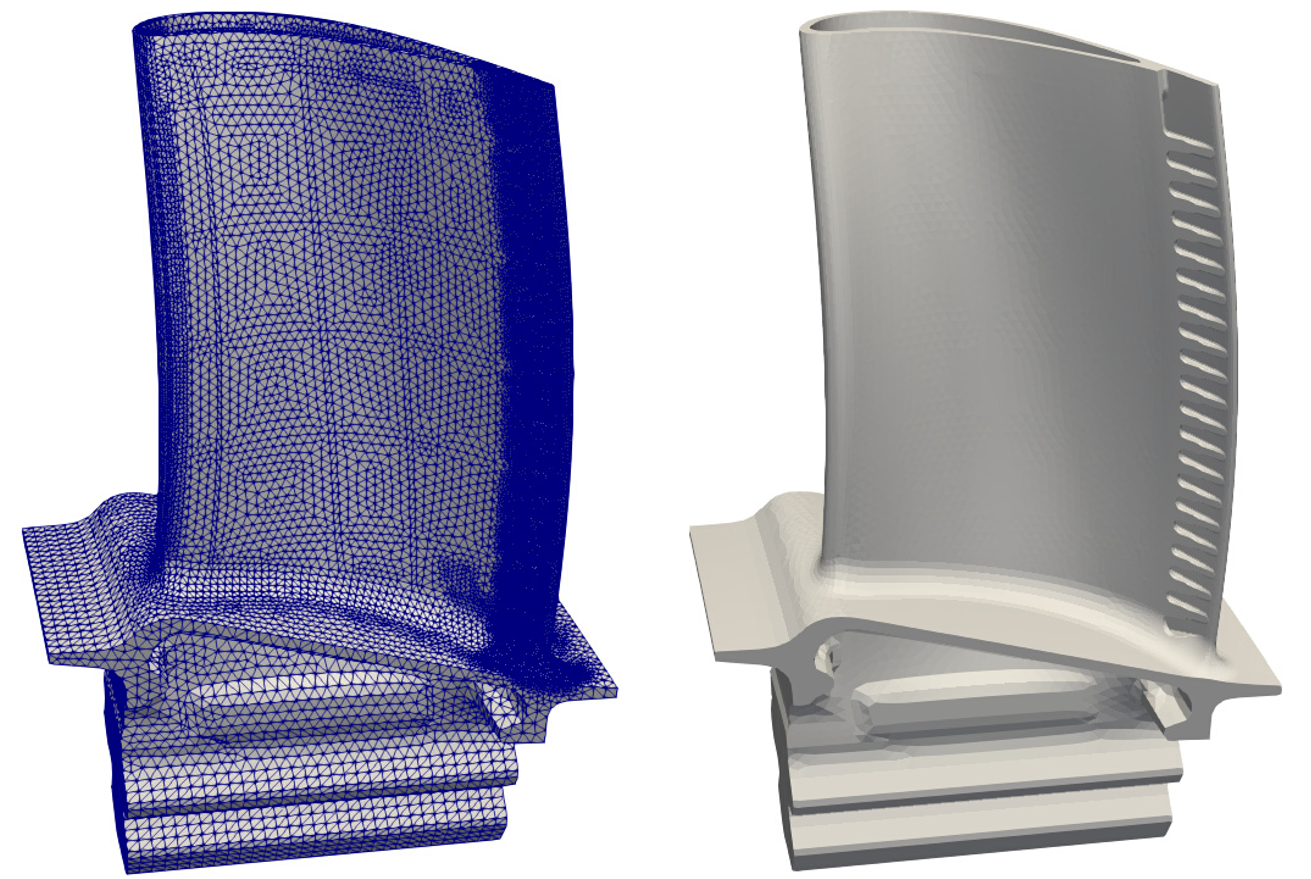}
\caption{High-pressure turbine blade geometry and mesh (micro-perforations are not modeled)}
\label{HPTurbBladeGeomMesh}
\end{figure}

Figure~\ref{HPTurbBladeGeomMesh} gives the geometry and the finite-element mesh of a real high-pressure turbine blade. The mesh is made of quadratic tetrahedral elements, and contains a number of nodes in the order of the million. The elasto-viscoplastic mechanical behavior is described by a crystal plasticity model presented in Section~\ref{SectionMechanicalModel}.
As explained above, Monte Carlo simulations using a commercial software as Abaqus are unaffordable.
With the help of domain decomposition methods, the computation time can be reduced by solving equilibrium equations in parallel on different subdomains of the geometry. Using the implementation of the Adaptive MultiPreconditioned FETI solver~\cite{feti} in Z-set finite-element software~\cite{zset}, the simulation of one single loading cycle of the HP turbine blade with 48 subdomains takes approximately 53 minutes.

\subsubsection{Objectives}
The objective is to use a ROM-net to quantify uncertainties on the mechanical behavior of the high-pressure turbine blade, given uncertainties on the thermal loading. The reduction of the computation time should enable Monte Carlo simulations for uncertainty quantification. 
In particular, we are not interested in predicting the state of the structure after a large number of flight-representative loading cycles. Only one cycle is simulated. Cyclic extrapolation of the behavior of a high-pressure turbine blade has been studied in~\cite{CasenaveIJNME} and is out of the scope of the present work.

\subsection{Modeling assumptions}

\subsubsection{Weak thermomechanical coupling}

It is assumed that the heat produced or dissipated by mechanical phenomena has negligible effects in comparison with thermal conduction, which enables avoiding strongly coupled thermomechanical simulations and running thermal and mechanical simulations separately instead. Under a weak thermomechanical coupling, the first step consists in solving the heat equation to determine the temperature field and its evolution over time. The temperature field history defines the thermal loading and is used to compute thermal strains and temperature-dependent material parameters for the mechanical constitutive laws. Once the thermal loading is known, the temperature-dependent mechanical problem must be solved in order to predict the mechanical response of the structure.

\subsubsection{Cyclic thermomechanical loading}

\begin{figure}[!h]
\centering
\includegraphics[scale=0.3]{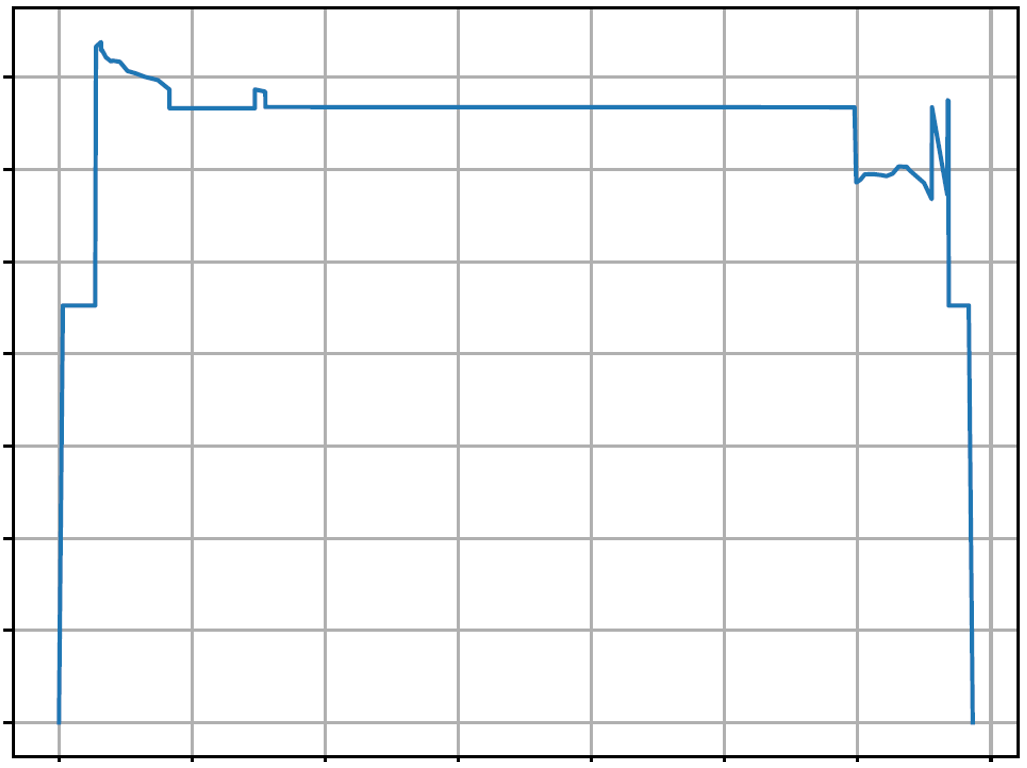}
\caption{Function $\omega(t)$ defining one cycle for the rotation speed.}
\label{RotationSpeed}
\end{figure}

The thermomechanical loading applied to the high-pressure turbine blade during its whole life is modeled as a cyclic loading, with one cycle being equivalent to one flight. The rotation speed of the turbine's rotor is proportional to a periodic function of time $\omega (t)$ whose evolution over one period (or cycle, see Figure~\ref{RotationSpeed}) is representative of one flight with its three main regimes, namely take-off, cruise, and landing. The period (or duration of one cycle) is denoted by $t_c$. The rotation speed between flights $k$ and $k+1$ is zero, which means that $\omega (k t_c) = 0$ for any integer $k$. The rotation speed $\omega (t)$ is scaled so that its maximum is $1$.

Let $\Omega \subset \mathbb{R}^3$ denote the solid body representing the high-pressure turbine blade, with $\partial \Omega$ denoting its outer surface. Let $\partial \Omega^p \subset \partial \Omega$ be the surface corresponding to the intrados and extrados. The thermal loading is defined as:\begin{equation}
\forall \bs{\xi} \in \Omega, \quad \forall t \in \mathbb{R}_{+}, \qquad T(\bs{\xi}, t) = (1 - \omega(t))T_{0} + \omega(t) T_{\max}(\bs{\xi})
\label{ThermalLoadingDef}
\end{equation} where $T_{0} = 293 \ K$ and $T_{\max}$ is the temperature field obtained when the rotation speed reaches its maximum. This field $T_{\max}$ is obtained either by an aerothermal simulation or by a stochastic model, as explained later. Similarly, the pressure load applied on $\partial \Omega^p$ reads:\begin{equation}
\forall \bs{\xi} \in \partial\Omega^p , \quad \forall t \in \mathbb{R}_{+}, \qquad p^{\partial\Omega}(\bs{\xi}, t) = (1 - \omega(t))p^{\partial\Omega}_{0} + \omega(t) p^{\partial\Omega}_{\max}(\bs{\xi})
\end{equation} where $p^{\partial\Omega}_{0} = 1 \ \textrm{atm}$ is the atmospheric pressure at sea level, and where $p^{\partial\Omega}_{\max}$ is the pressure field obtained when the rotation speed reaches its maximum. The clamping of the blade's fir-tree foot on the rotor disk is modeled by displacements boundary conditions that are not detailed here.

\subsubsection{Geometric details and thermal barrier coating}

Small geometric details of the structure have been removed to simplify the geometry. Nonetheless, the main cooling channels are considered. The effects of the thermal barrier coating (TBC) have been integrated in aerothermal simulations, but the TBC is not considered in the mechanical simulation although its damage locally increases the temperature in the nickel-based superalloy and thus affects the fatigue resistance of the structure. Additional centrifugal effects due to the TBC are not taken into account.

\subsubsection{Influential factors}

The predicted mechanical response of the structure depends on many different factors. Below is a nonexhaustive list of influential factors that are possible sources of uncertainties in the numerical simulation:
\begin{itemize}
\item \textbf{Thermal loading:} The viscoplastic behavior of the nickel-based superalloy is very sensitive to the temperature field and its gradients. However, the temperature field is not accurately known because of the impossibility of validating numerical predictions experimentally. Indeed, temperature-sensitive paints are accurate to within $50 \ K$ only, and they do not capture a real surface temperature field since they measure the maximum temperature reached locally during the experiment.
\item \textbf{Crystal orientation:} Because of the complexity of the manufacturing process of monocrystalline blades, the orientation of the crystal is not perfectly controlled. As the superalloy has anisotropic mechanical properties, defaults in crystal orientation highly affect the location of damaged zones in the structure.
\item \textbf{Mechanical loading:} The centrifugal forces are well known because they are related to the rotation speed that is easy to measure. On the contrary, pressure loads are uncertain because of the turbulent nature of the incoming fluid flow. However, the effects of pressure loads uncertainties on the mechanical response are less significant than those of the thermal loading and crystal orientation uncertainties.
\item \textbf{Constitutive laws:} Uncertainties on the choice of the constitutive model, the relevance of the modeling assumptions, and the values of the calibrated parameters involved in the constitutive equations also influence the results of the numerical simulations.
\end{itemize}

For simplification purposes, the only source of uncertainty that is considered in this work is the thermal loading. The equations of the mechanical problem are then seen as parametrized equations, where the parameter is the temperature field $T_{\max}$ (see Equation~\eqref{ThermalLoadingDef}) obtained when the rotation speed reaches its maximum value. The dimension of the parameter space is then the number of nodes in the finite-element mesh. The mechanical loading is assumed to be deterministic. With the crystal orientation, the constitutive laws and their parameters (or coefficients), they are considered as known data describing the context of the study and given by experts.

\subsection{Stochastic model for the thermal loading}

\begin{figure}[!h]
\centering
\includegraphics[scale=0.35]{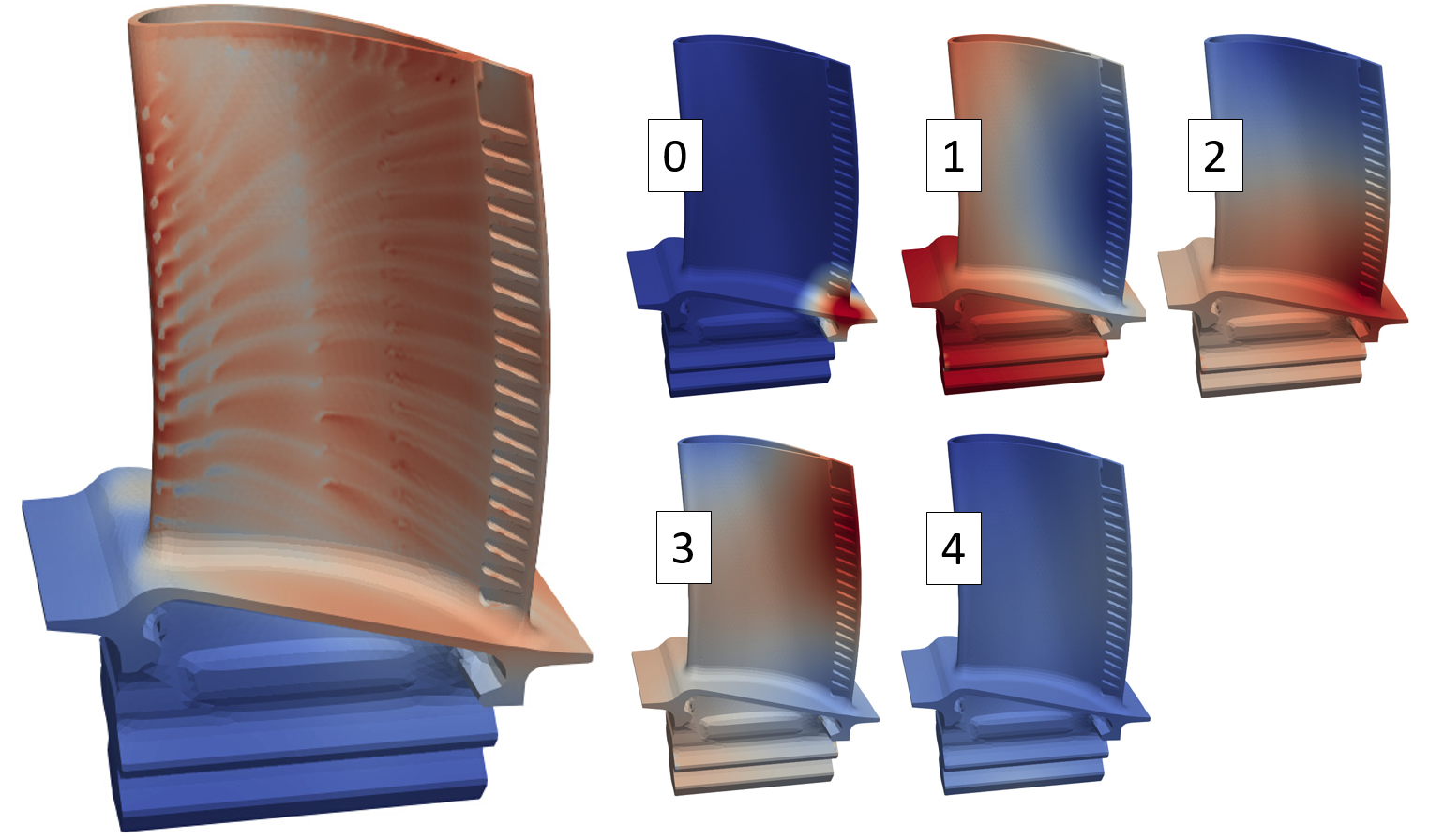}
\caption{Reference temperature field (on the left), temperature perturbation at the trailing edge (field 0 = $\delta T_0$), and fluctuation modes (fields 1 to 4). The fluctuations in the fourth mode are located inside the blade, in the cooling channels.}
\label{ThermalLoadingModel}
\end{figure}

A stochastic model is required to take into account the uncertainties on the thermal loading. Given the definition of the thermal loading in Equation~\eqref{ThermalLoadingDef}, we only need to model uncertainties in space through the field $T_{\max}$ obtained when the rotation speed reaches its maximum value. The random temperature fields must satisfy some constraints: they must satisfy the heat equation, and they must not take values out of the interval $[ 0 \ K ; T_{\textrm{melt}} ]$, where $T_{\textrm{melt}}$ is the melting point of the superalloy. These random fields are obtained by adding random fluctuations to a reference temperature field, see Figure~\ref{ThermalLoadingModel}. The reference field and comes from aerothermal simulations run with the software  \textit{Ansys Fluent}\footnote{\url{https://www.ansys.com/products/fluids/ansys-fluent}}. The data-generating distribution is defined as a Gaussian mixture model made of two Gaussian distributions with the same covariance function but with distinct means, and with a prior probability of $0.5$ for each Gaussian distribution. The Gaussian distributions are obtained by taking the four first eigenfunctions of the covariance function (see Karhunen-Lo\`{e}ve expansion~\cite{10.1007/s00607-008-0018-3}), with a standard deviation of $15 \ K$. Therefore, realizations of the random temperature field read:\begin{equation}
\forall \bs{\xi} \in \Omega, \quad T(\bs{\xi}) = T_{\textrm{ref}}(\bs{\xi}) + \Upsilon_{0} \ \delta T_0 (\bs{\xi}) + \sum_{i=1}^{4} \Upsilon_{i} \ \delta T_i (\bs{\xi})
\label{EquationDefiningRandTempField}
\end{equation} where $T_{\textrm{ref}}$ is the reference field, $\delta T_0 $ is a temperature perturbation at the trailing edge whose maximum value is $50 \ K$, $\{ \delta T_i \}_{1 \leq i \leq 4}$ are fluctuation modes, $\Upsilon_{0}$ is a random variable following the Bernoulli distribution with parameter $0.5$, and $\{ \Upsilon_i \}_{1 \leq i \leq 4}$ are independent and identically distributed random variables following the standard normal distribution $\mathcal{N}(0,1)$. The variable $\Upsilon_{0}$ is also independent of the other variables $\Upsilon_i$. The different fields involved in Equation~\eqref{EquationDefiningRandTempField} can be visualized in Figure~\ref{ThermalLoadingModel}. Equation~\eqref{EquationDefiningRandTempField} defines a mixture distribution with two Gaussian distributions whose means are $T_{\textrm{ref}}$ and $T_{\textrm{ref}} + \delta T_0$. We voluntarily define this mixture distribution with $\delta T_0$ adding $50 \ K$ in a critical zone of the turbine blade in order to check that our physics-informed cluster analysis can successfully detect two relevant clusters, \textit{i.e.} one for fields obtained with $\Upsilon_{0}(\theta) = 0$ and one for fields obtained with $\Upsilon_{0}(\theta) = 1$. Indeed, the temperature perturbation $\delta T_0$ is expected to significantly modify the mechanical response of the high-pressure turbine blade. All the fields $\{ \delta T_i \}_{0 \leq i \leq 4}$ satisfy the steady heat equation like $T_{\textrm{ref}}$, which ensures that the random fields always satisfy the heat equation under the assumption of a linear thermal behavior. For nonlinear thermal behaviors, Equation~\eqref{EquationDefiningRandTempField} would define surface temperature fields that would be used as Dirichlet boundary conditions for the computation of bulk temperature fields. The assumption of a linear thermal behavior is adopted here to avoid solving the heat equation for every realization of the random temperature field.

Let us now give more details about the construction of the fluctuation modes $\{ \delta T_i \}_{1 \leq i \leq 4}$. First, surface fluctuation modes are computed on the boundary $\partial\Omega$ using the method given in~\cite{RF_on_curved_surf} for the construction of random fields on a curved surface. The correlation function is defined as a function of the geodesic distance $d_G$ along the surface $\partial\Omega$:\begin{equation}
\rho(\bs{\xi},\bs{\xi}') = \exp\left( -\frac{d_{G}(\bs{\xi},\bs{\xi}')}{d_{G}^0} \right)
\end{equation} where $d_{G}^0$ is a correlation length. Geodesic distances are computed thanks to the algorithm described in~\cite{MMP87, MMP87implemented} and implemented in the Python library \textit{gdist}\footnote{\url{https://pypi.org/project/gdist/}}. A covariance matrix is built by evaluating the correlation function on pairs of nodes of the outer surface of the finite-element mesh, and multiplying the correlation by the constant variance. The four surface modes are then obtained by finding the four eigenvectors corresponding to the largest eigenvalues of the covariance matrix. The steady heat equation with Dirichlet boundary conditions is solved for each of these surface modes to derive the 3D fluctuation modes, using \textit{Z-set}~\cite{zset} finite-element solver. The Python library \textit{BasicTools}\footnote{\url{https://gitlab.com/drti/basic-tools}} developed by SafranTech is used to read the finite-element mesh and write the temperature fields in a format that can be used for simulations on \textit{Z-set}.

\subsection{Mechanical constitutive model}
\label{SectionMechanicalModel}

It is assumed that the mechanical behavior of the high-pressure turbine blade can be described in the framework of the infinitesimal strain theory. The mechanical response of the structure during the first loading cycle is described by the following equilibrium equations and boundary conditions:\begin{equation}
\left\lbrace
\begin{array}{crclcl}
  & \g{div}(\bs{\sigma}(\bs{\xi},t)) + f_{C}(\bs{\xi},t) & = &\g{0} & \forall t\in[0;t_c] & \ \  \forall \bs{\xi}\in\Omega \\
& \bs{\sigma}(\bs{\xi},t).\g{n}(\bs{\xi},t) & = & - p^{\partial\Omega}(\bs{\xi}, t) \g{n}(\bs{\xi},t) & \forall t\in[0;t_c] & \ \  \forall\bs{\xi}\in \partial \Omega^p \\
& \g{u}(\bs{\xi},t) & = & \g{u}^{\partial \Omega}(\bs{\xi},t)  & \forall t\in[0;t_c] & \ \  \forall\bs{\xi}\in \partial \Omega \setminus \partial\Omega^p \,
\end{array}\right.
\label{equilibriumEqs}
\end{equation} where $\g{u}(\bs{\xi},t)$ is the displacement field (primal variable), $\bs{\sigma}(\bs{\xi},t)$ is the symmetric second-order Cauchy stress tensor, $f_{C}(\bs{\xi},t)$ is the local volumic centrifugal force, $\g{u}^{\partial \Omega}(\bs{\xi},t)$ are the imposed displacements, and $\g{n}(\bs{\xi},t)$ is the outward-pointing normal vector to the outer surface $\partial\Omega$. The relation between the stress tensor and the displacement field is described by constitutive laws modeling the mechanical behavior of the monocrystalline nickel-based superalloy. At high temperatures, this material has an elasto-viscoplastic behavior that can be described in the crystal plasticity framework~\cite{10.1115/1.3167205, 10.1115/1.2903374} to model inelastic strains generated by the motion of dislocations\footnote{Linear defects in the crystal structure.} in different slip systems of the crystal. The strain tensor $\varepsilon$ is defined as the symmetric part of the displacement gradient (with respect to $\bs{\xi}$):\begin{equation}
\bs{\varepsilon} = \frac{1}{2}\left( \bs{\nabla}\g{u} + (\bs{\nabla}\g{u})^T \right)
\end{equation}

\begin{figure}[!h]
\centering
\includegraphics[scale=0.35]{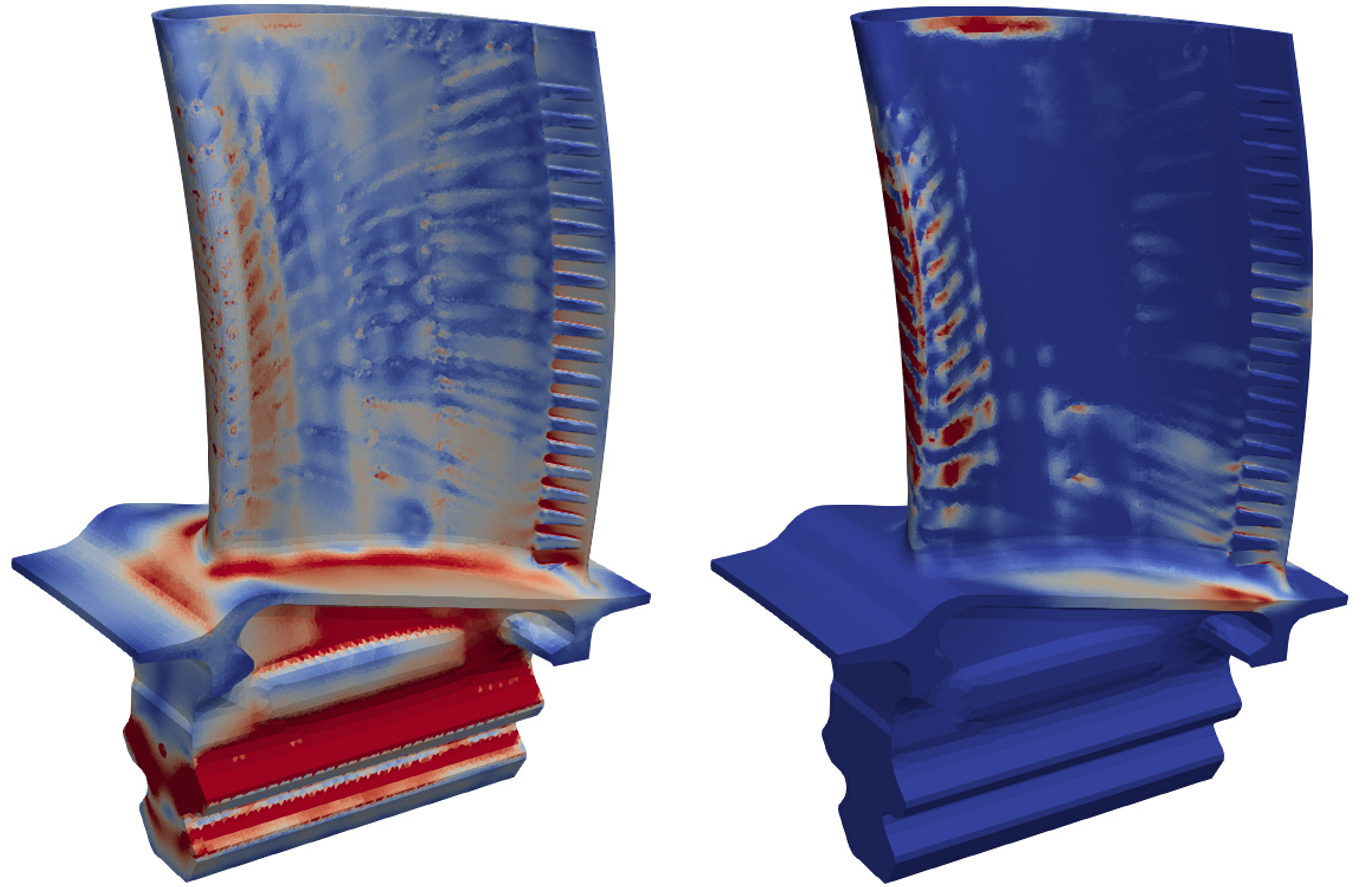}
\caption{On the left: von Mises stress field $\sigma_{\textrm{eq}}$ obtained when the rotation speed reaches its maximum value. On the right: accumulated plastic strain $p_{\textrm{cum}}^{o}$ in octahedral slip systems at the end of the first cycle. Note: the foot of the high-pressure turbine blade has an elastic behavior, while the rest of the blade has a viscoplastic behavior described by a crystal plasticity model.}
\label{QoIRef}
\end{figure}

The stress tensor is obtained from the elastic strain tensor thanks to Hooke's law: \begin{equation}
\bs{\sigma} = \g{C} :  \left( \bs{\varepsilon} - \bs{\varepsilon}^p - \alpha (T - T_0) \g{1} \right)
\end{equation} where $\bs{\varepsilon}^p$ is the tensor of inelastic strains and $\g{1}$ is the identity second-order tensor. The fourth-order tensor $\g{C}$ is the stiffness tensor. Given the face-centered cubic crystal structure of the superalloy, the stiffness tensor is anisotropic but has only three independent coefficients. The thermal expansion of crystals with cubic symmetry is isotropic, which explains why the thermal expansion coefficient $\alpha$ is the same in all directions. The time evolution of hidden variables such as inelastic strains are described by ordinary differential equations that must be solved at every integration point of the finite-element mesh. The inelastic strain rate can be decomposed into contributions of dislocations motions in 12 octahedral slip systems and 6 cubic slip systems:\begin{equation}
\dot{\bs{\varepsilon}}^p = \sum_{s=1}^{12} \dot{\gamma}_{s}^{o} \  \text{sym} \left( \g{l}_{s}^{o} \otimes \g{n}_{s}^{o} \right) + \sum_{s=1}^{6} \dot{\gamma}_{s}^{c} \  \text{sym} \left( \g{l}_{s}^{c} \otimes \g{n}_{s}^{c} \right) = \sum_{s=1}^{12} \dot{\gamma}_{s}^{o} \g{m}_{s}^{o} + \sum_{s=1}^{6} \dot{\gamma}_{s}^{c} \g{m}_{s}^{c}
\end{equation} where $\dot{\gamma}_{s}^{o}$ (\textit{resp.} $\dot{\gamma}_{s}^{c}$) is the shear strain rate in the $s$-th octahedral (\textit{resp.} cubic) slip system. The tensor $\g{m}_{s}^{o}$ (\textit{resp.} $\g{m}_{s}^{c}$) is the orientation tensor of the $s$-th octahedral (\textit{resp.} cubic) slip system, defined by the normal $\g{n}_{s}^{o}$ (\textit{resp.} $\g{n}_{s}^{c}$) to the slip plane and the slip direction $\g{l}_{s}^{o}$ (\textit{resp.} $\g{l}_{s}^{c}$). The shear strain rates $\dot{\gamma}_{s}^{o}$ are given by a hyperbolic viscoplastic flow rule:\begin{equation}
\dot{\gamma}_{s}^{o} = \varepsilon_{h}^{o} \ \textrm{sinh} \left( \left\langle \frac{| \tau_{s}^{o} - x_{s}^{o} | - r_{s}^{o}}{K_{h}^{o}} \right\rangle ^{n_{h}^{o}} \right) \ \textrm{sign}(\tau_{s}^{o} - x_{s}^{o})
\end{equation} where $\varepsilon_{h}^{o}$, $K_{h}^{o}$ and $n_{h}^{o}$ are material parameters. Similar equations are satisfied in cubic slip systems. The resolved shear stresses $\tau_{s}^{o}$ are given by Schmid's law:\begin{equation}
\tau_{s}^{o} = \bs{\sigma} : \g{m}_{s}^{o} 
\end{equation} Again, similar equations are valid for cubic slip systems. The stress variables $x_{s}^{o}$, $x_{s}^{c}$, $r_{s}^{o}$ and $r_{s}^{c}$ describe hardening phenoma, \textit{i.e.} the evolution of the shape of the elastic domain within which no dissipative phenoma occur. The back-stresses $x_{s}^{o}$ (and $x_{s}^{c}$) are the solutions of an ordinary differential equation modeling kinematic hardening with static recovery:\begin{equation}
\dot{x}_{s}^{o} = c^{o} \dot{\gamma}_{s}^{o} - d^{o} x_{s}^{o} | \dot{\gamma}_{s}^{o} | - c^{o} \left( \frac{| x_{s}^{o} |}{M^o} \right)^{m^o}
\end{equation} Isotropic hardening is modeled by the following equations:\begin{equation}
r_{s}^{o} = r_{0}^{o} + Q^{o} \left( 1 - \exp \left( - b^{o} \nu_{s}^{o} \right) \right)
\end{equation} with $\dot{\nu}_{s}^{o} = | \dot{\gamma}_{s}^{o} |$. All the constitutive equations given in this section are true for all $\bs{\xi}\in\Omega$ and for all $t \in [0;t_c]$, and are solved at every integration point of the finite-element mesh. All the coefficients involved in these equations depend on the local value of the temperature field. The problem is thus seen as a system of partial differential equations and ordinary differential equations parametrized by the thermal loading. The standard procedure for the computation of a fatigue lifetime with an uncoupled damage model consists in solving the mechanical problem for a large number of cycles until the stabilization of the mechanical response (in the case of plastic shakedown). Then, a damage field can be computed in a post-processing step and can be linked to a fatigue lifetime. For high-pressure turbine blades, fatigue models generally consider interaction effects with oxidation and creep, like in~\cite{gallerneau1995etude, FatOxFlu}. In this work, no fatigue lifetime is computed since we only solve the problem for the very first cycle. Instead, our quantity of interest is a strain indicator that partially describes the damage state of the material. This quantity of interest corresponds to the accumulated plastic strain in octahedral slip systems at the end of the first cycle, which reads:\begin{equation}
p_{\textrm{cum}}^{o}(\bs{\xi}) = \int_{0}^{t_c} \sqrt{\frac{2}{3} \  \dot{\bs{\varepsilon}}^{p,o}(\bs{\xi},t) : \dot{\bs{\varepsilon}}^{p,o}(\bs{\xi},t)} \ dt
\end{equation} with:\begin{equation}
\dot{\bs{\varepsilon}}^{p,o} = \sum_{s=1}^{12} \dot{\gamma}_{s}^{o} \g{m}_{s}^{o}
\end{equation} It is also common to look at the values of the von Mises equivalent stress field defined as:\begin{equation}
\sigma_{\textrm{eq}} = \sqrt{\frac{3}{2}  \ \g{s}:\g{s}},  \qquad   \g{s} = \bs{\sigma} - \frac{1}{3} \textrm{tr}(\bs{\sigma})\g{1}
\end{equation} Therefore, the variables considered for the evaluation of the ROM-net and for uncertainty quantification are the accumulated plastic strain $p_{\textrm{cum}}^{o}$ in octahedral slip systems at the end of the first cycle, and the von Mises stress $\sigma_{\textrm{eq}}$ obtained when the rotation speed reaches its maximum value. These variables can be visualized in Figure~\ref{QoIRef} for a reference thermal loading.

\section{ROM-net based uncertainty quantification applied to an industrial high-pressure turbine blade}
\label{sec:indusStudy}

This section develops the different stages of the ROM-net for the industrial test case presented in the previous section. Given our budget of 200 high-fidelity simulations, a dictionary containing two local ROMs is constructed thanks to a physics-informed clustering procedure. A logistic regression classifier is trained for automatic model recommendation using information identified by feature selection, followed by an alternative to the Gappy-POD for full-field reconstruction of dual quantities. Then, the results of the uncertainty quantification procedure are presented. We advise the reader to refer to Section~\ref{sec:worfklow}, which contains an illustration of the proposed workflow, while reading Sections~\ref{sec:does}-\ref{sec:UQres}. Finally the accuracy of the ROM-net is validated using simulations for new temperature loadings.

\subsection{Design of numerical experiments}
\label{sec:does}

Given the computational cost of high-fidelity mechanical simulations of the high-pressure turbine blade, the training data are sampled from the stochastic model for the thermal loading using a design of experiments (DoE). Our computational budget corresponds to 200 high-fidelity simulations, so a database of 200 temperature fields must be built. This database includes two separate datasets coming from two independent DoEs:
\begin{itemize}
\item The first dataset is built from a Maximum Projection LHS design (\textit{MaxProj LHS DoE}) and contains 80 points. This dataset will be used for the construction of the dictionary of local ROMs via clustering. The MaxProj LHS DoE has good space-filling properties on projections onto subspaces of any dimension.
\item The second dataset is built from a Sobol' sequence (\textit{Sobol' DoE}) of 120 points. Using a suboptimal DoE method ensures that this second dataset is different and independent from the first one. The lower quality of this dataset with respect to the first one is compensated by its larger population. This dataset will be used for learning tasks requiring more training examples than the construction of the local ROMs, namely the classification task for automatic model recommendation, and the training of cluster-specific surrogate models for the reconstruction of full fields from hyper-reduced predictions on a reduced-integration domain. These surrogate models (\textit{Gappy surrogates}) replace the Gappy-POD~\cite{GappyPOD} method that is commonly used in hyper-reduced simulations to retrieve dual variables on the whole mesh.
\end{itemize}

\begin{figure}[!h]
\centering
\includegraphics[scale=0.45]{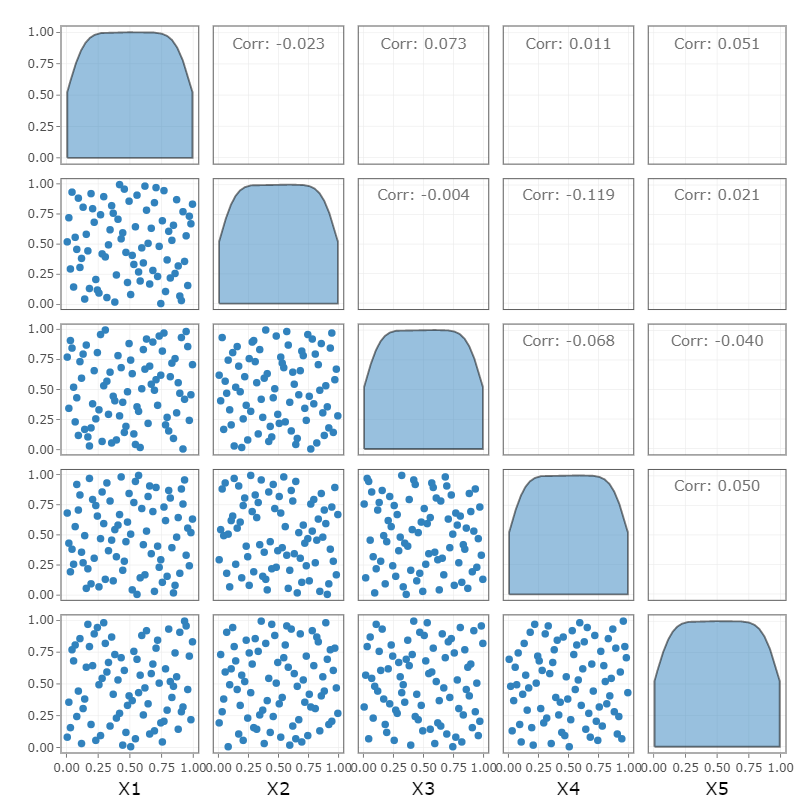}
\caption{Visualization of the MaxProj LHS DoE. The marginal distributions are represented on the diagonal. The 5D DoE is projected on 2D subspaces for visualization purposes, in order to check space-filling properties in 2D.}
\label{MaxProjDOE}
\end{figure}

\begin{figure}[!h]
\centering
\includegraphics[scale=0.45]{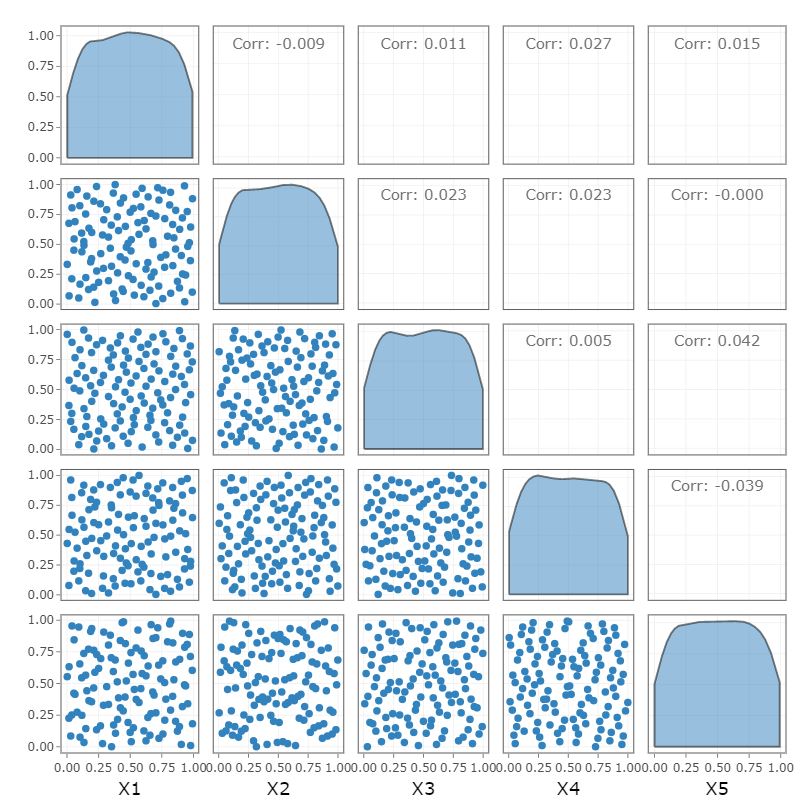}
\caption{Visualization of the Sobol' DoE. The marginal distributions are represented on the diagonal. The 5D DoE is projected on 2D subspaces for visualization purposes, in order to check space-filling properties in 2D.}
\label{SobolDOE}
\end{figure}

These DoEs are built with the platform \textit{Lagun}\footnote{\url{https://gitlab.com/drti/lagun}}. The fact that these two datasets come from two separate DoEs is beneficial: as each of them is supposed to have good space-filling properties, they are both representative of the possible thermal loading and can therefore be used to define a training set and a test set for a given learning task. For instance, the classifier trained on the Sobol' DoE can be tested on the MaxProj LHS DoE. The local ROMs built from snapshots belonging to the MaxProj LHS DoE can make predictions on the Sobol' DoE that will be used for the training of the Gappy surrogates, which is relevant since the Gappy surrogates are supposed to analyze ROM predictions on new unseen data in the exploitation phase.

Drawing random temperature fields as defined in Equation~\eqref{EquationDefiningRandTempField} requires sampling data from the random variables $\{ \Upsilon_i \}_{0\leq i \leq 4}$, where $\Upsilon_0$ follows the Bernoulli distribution with parameter $0.5$ and the variables $\Upsilon_i$ for $i \in [\![ 1;4 ]\!]$ are independent standard normal variables and independent of $\Upsilon_0$. Both DoE methods (Maximum Projection LHS and Sobol' sequence) generate point clouds with a uniform distribution in the unit hypercube. Figures~\ref{MaxProjDOE} and~\ref{SobolDOE} show the projections onto 2-dimensional subspaces of the 5D point clouds used to build our datasets. The marginal distributions are plotted to check that they well approximate the uniform distribution. These point clouds, considered as samples of a random vector $(\chi_0, \chi_1, \chi_2, \chi_3, \chi_4)$ following the uniform distribution on the unit hypercube, are transformed into realizations of the random vector $(\Upsilon_0, \Upsilon_1, \Upsilon_2, \Upsilon_3, \Upsilon_4)$ using the following transformations:\begin{equation}
\Upsilon_0 = \mathds{1}_{\chi_0 > 1/2} \qquad \textrm{and} \quad \forall i \in [\![1;4]\!], \quad \Upsilon_i = F^{-1}(\chi_i)
\label{TransfoUniformToTempReducedCoords}
\end{equation} where $F^{-1}$ is the inverse of the cumulative distribution function of the standard normal distribution. The resulting samples define the MaxProj dataset and the Sobol' dataset of random temperature fields, using Equation~\eqref{EquationDefiningRandTempField}. Each temperature field defines a thermal loading, using Equation~\eqref{ThermalLoadingDef}. The 200 corresponding mechanical problems are solved for one loading cycle with the finite-element software \textit{Z-set}~\cite{zset} with the domain decomposition method described in~\cite{feti}, with 48 subdomains. The average computation time for one simulation is 53 minutes.

\subsection{ROM dictionary construction}

\subsubsection{Clustering}

\begin{figure}[!h]
\centering
\includegraphics[scale=0.32]{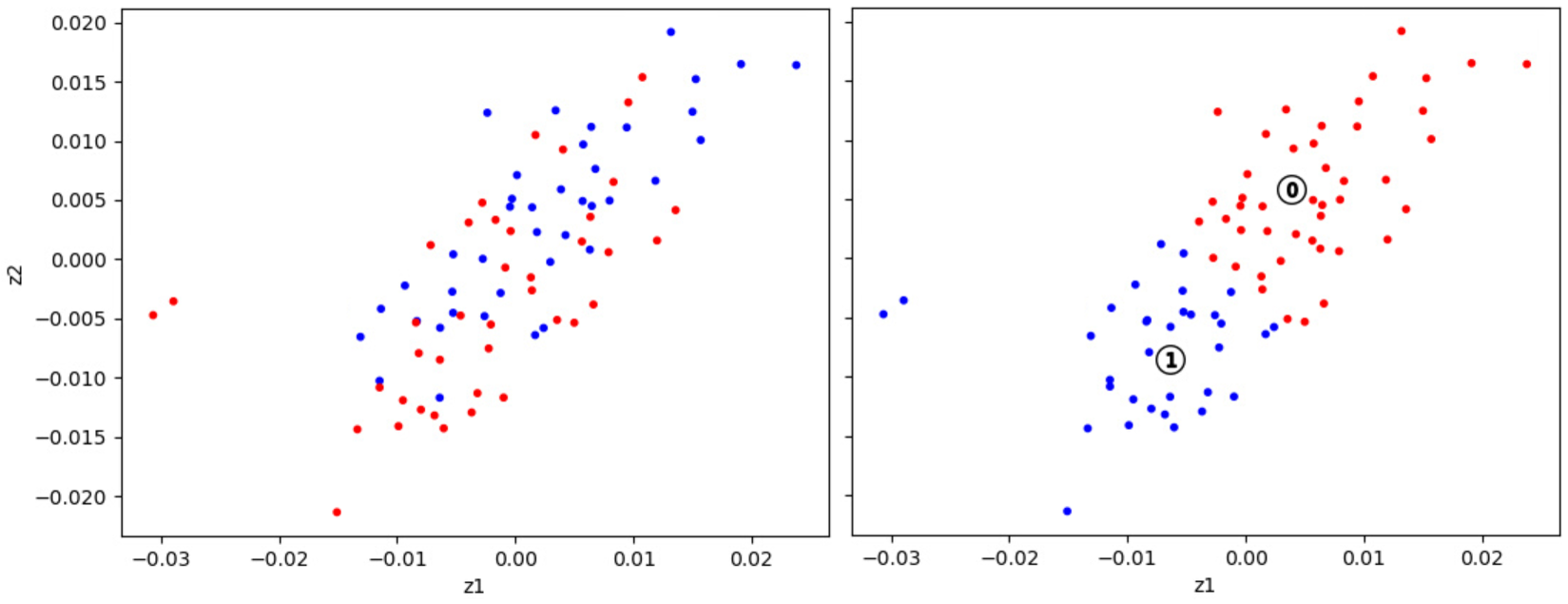}
\caption{MDS representation of the clustering results using the ROM-oriented dissimilarity measure on the displacement field (method-oriented variant). On the left, the colors correspond to the expected clusters. On the right, the colors correspond to the clusters identified by the clustering algorithm. The positions of the labels $0$ and $1$ coincide with the positions of the clusters' medoids. The MDS relative error $\varsigma(\g{Z}_{0};\bs{\delta})/\varsigma(\g{0};\bs{\delta})$ is $7.9\%$.}
\label{ClusteringResultsMDSu}
\end{figure}

\begin{figure}[!h]
\centering
\includegraphics[scale=0.3]{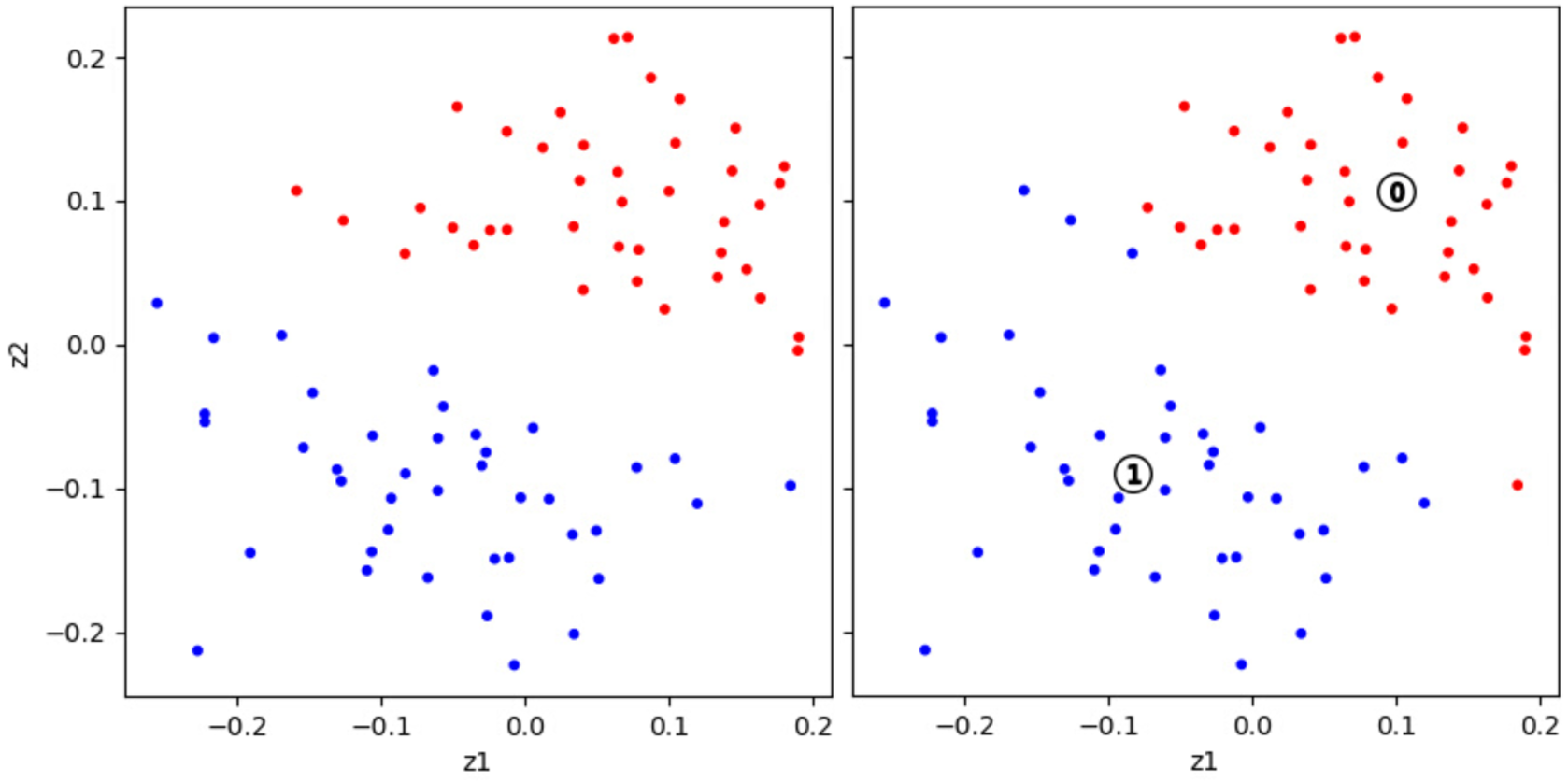}
\caption{MDS representation of the clustering results using the ROM-oriented dissimilarity measure on the quantity of interest $p_{\textrm{cum}}^{o}$ (goal-oriented variant). On the left, the colors correspond to the expected clusters. On the right, the colors correspond to the clusters identified by the clustering algorithm. The positions of the labels $0$ and $1$ coincide with the positions of the clusters' medoids. The MDS relative error $\varsigma(\g{Z}_{0};\bs{\delta})/\varsigma(\g{0};\bs{\delta})$ is $12\%$.}
\label{ClusteringResultsMDSevgeq}
\end{figure}

\begin{figure}[!h]
\centering
\includegraphics[scale=0.35]{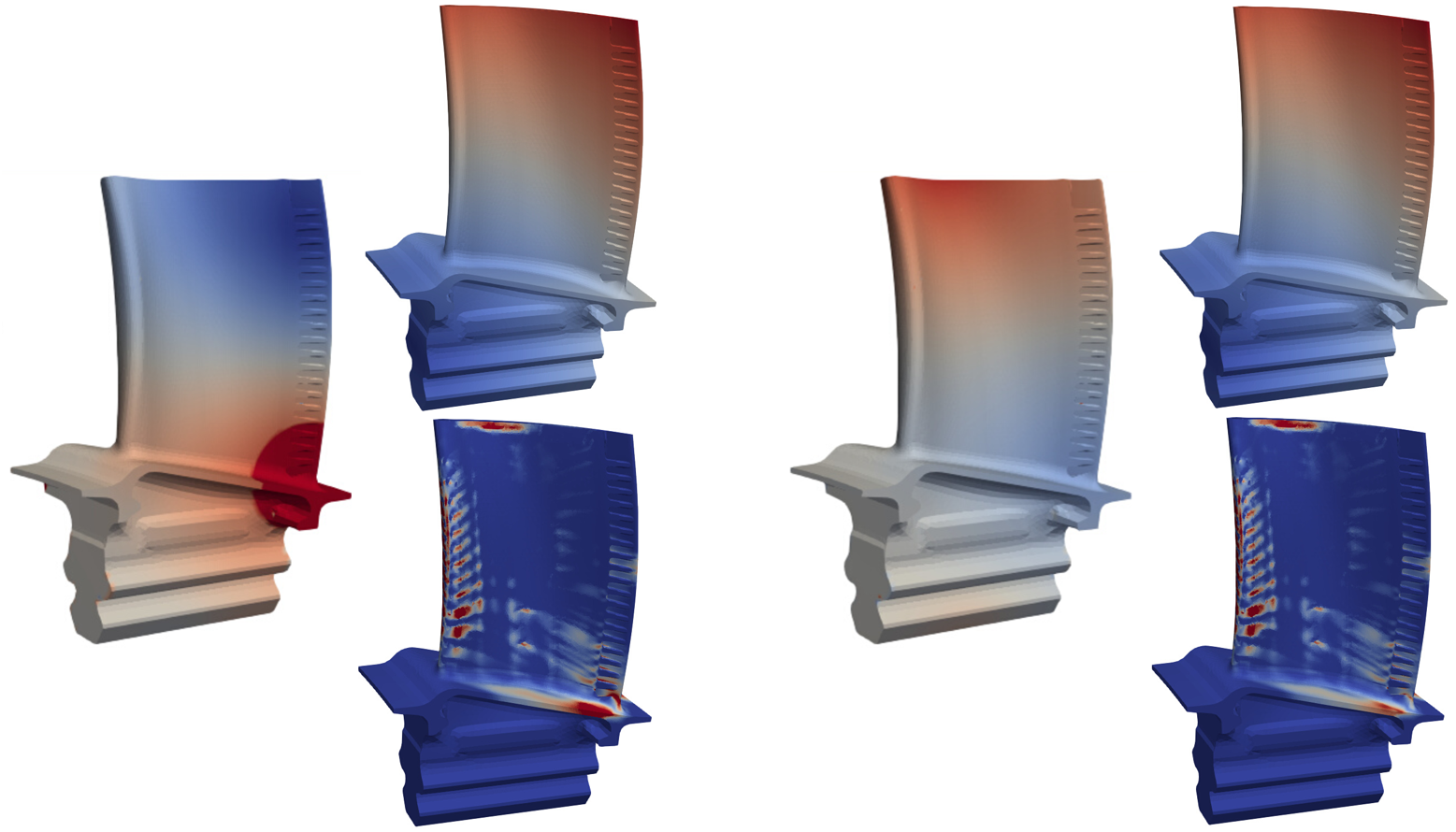}
\caption{The 3 fields on the left correspond to the medoid of cluster $0$, and those on the right correspond to the medoid of cluster $1$. The fields in the first and the third columns show the differences between the medoids' temperature fields and the reference temperature field $T_{\textrm{ref}}$ (the scale is truncated for the first field). The second and the fourth columns show the displacement magnitude field $\sqrt{\g{u}.\g{u}}$ (top) and the quantity of interest $p_{\textrm{cum}}^{o}$ (bottom).}
\label{MedoidsFields}
\end{figure}

The 80 simulations associated to the MaxProj dataset are used as clustering data. Loading all the simulation data and computing the pairwise ROM-oriented dissimilarities takes about 5 minutes. The ROM-oriented dissimilarity defined in~\cite[Definition~3.11]{daniel2021physicsinformed}, and mentioned in Section~\ref{sec:dicROMnets}, is computed with $n=1$, \textit{i.e.} each simulation is represented by one field. Two variants are tested: a method-oriented variant, where the dissimilarities are computed from the displacements fields at the maximum rotation speed, and a goal-oriented variant, where the dissimilarities involve the quantity of interest $p_{\textrm{cum}}^{o}$ (accumulated plastic strain in octahedral slip systems at the end of the simulation). The dataset is partitioned into two clusters using our implementation of PAM~\cite{kMedoidsPAM, kaufman1990finding} k-medoids algorithm, with 10 different random initializations for the medoids. The clustering results can be visualized thanks to Multidimensional Scaling (MDS)~\cite{mds}. MDS is an information visualization method which consists in finding a low-dimensional dataset $\g{Z}_{0}$ whose matrix of Euclidean distances $\g{d}(\g{Z}_{0})$ is an approximation of the true dissimilarity matrix $\bs{\delta}$. To that end, a cost function called stress function is minimized with respect to $\g{Z}$:\begin{equation}
\g{Z}_{0} = \argmin_{\g{Z}} \left( \varsigma(\g{Z};\bs{\delta}) \right) = \argmin_{\g{Z}} \left( \sum_{i<j}(\delta_{ij}-d_{ij}(\g{Z}))^{2} \right)
\end{equation} This minimization problem is solved with the algorithm Scaling by MAjorizing a COmplicated Function (SMACOF, \cite{smacof}) implemented in Scikit-Learn~\cite{scikit-learn}. Figures~\ref{ClusteringResultsMDSu} and~\ref{ClusteringResultsMDSevgeq} show the clusters on the MDS representations with the two variants of the ROM-oriented dissimilarity measure. Each figure compares the clustering results with the expected clusters corresponding to $\Upsilon_0 = 0$ and $\Upsilon_0 = 1$, the latter corresponds to the perturbation $\delta T_0$ being activated. On this example, the method-oriented variant using the displacement field does not manage to distinguish the expected clusters. On the contrary, the goal-oriented variant leads to clusters that almost correspond to the expected ones, with only 4 points with wrong labels out of 80. In the sequel, the results obtained with the goal-oriented variant are considered. The medoids of the two clusters are given in Figure~\ref{MedoidsFields}. Cluster $0$ contains temperature fields for which $\Upsilon_0 = 1$, while cluster $1$ contains fields for which $\Upsilon_0 = 0$. It can be observed that the quantity of interest clearly differs from one cluster to the other, while the differences are hardly visible on the displacement field. The displacement field combines deformations associated to different phenomena (thermal expansion, elastic strains, viscoplastic strains) that are not necessarily related to damage in the structure, which could explain why the quantity of interest $p_{\textrm{cum}}^{o}$ seems to be more appropriate for clustering in this example.

\subsubsection{Construction of local ROMs}

\begin{figure}[!h]
\centering
\includegraphics[scale=0.8]{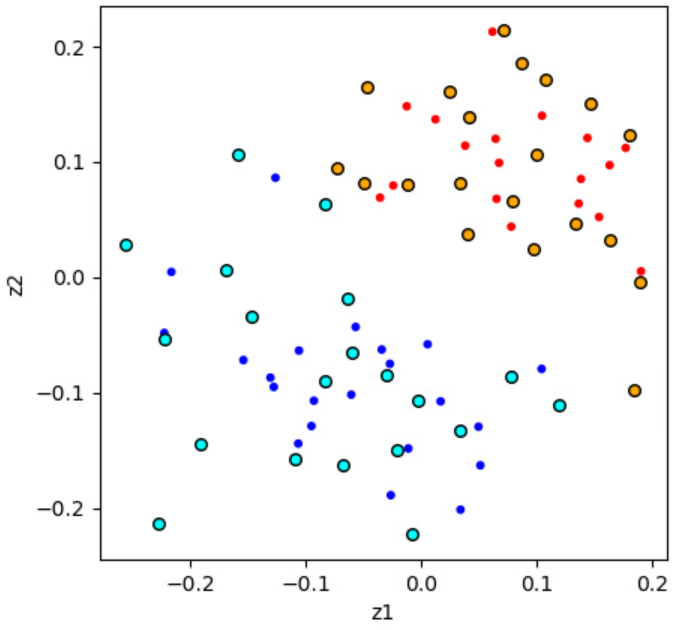}
\caption{MDS representation of the clustering results. Orange points represent the snapshots selected for cluster $0$, while the light blue points represent the snapshots selected for cluster $1$. For each cluster, the snapshots are selected by a maximin procedure starting from the medoid.}
\label{maximinSnapshotsSelection}
\end{figure}

The simulations used for the physics-informed clustering procedure can directly provide snapshots for the construction of the local ROMs. To control the duration of their training, only 20 simulations are selected to provide snapshots for the each local ROMs, which represents half of the clusters' populations. These simulations are selected in a maximin greedy approach starting from the medoid (see \cite[Algorithm~2,~Stage~2]{mca26010017} for a example of maximin selection). Figure~\ref{maximinSnapshotsSelection} shows which simulations have been selected for the construction of the local ROMs.

The local ROMs are built following the methodology described in Section~\ref{sec:nonlinearROM}, using the \textit{Mordicus} code developed in the FUI project MOR\_DICUS. The snapshot-POD and the ECM are done in parallel with shared memory on 24 cores. The tolerance for the snapshot-POD is set to $10^{-8}$ for the displacement field, and to $10^{-4}$ for dual variables (the quantity of interest $p_{\textrm{cum}}^{o}$ and the six components of the stress tensor). The POD bases for the dual variables will be used for their reconstruction with the Gappy surrogates. The tolerance for the ECM is set to $5 \times 10^{-4}$. The primal POD bases of both local ROMs contain 18 displacement modes. The local ROM $0$ (\textit{resp.} $1$) has 10 (\textit{resp.} 12) modes for the quantity of interest $p_{\textrm{cum}}^{o}$, and both ROMs have between 8 and 13 modes for stress components. The ECM selects 506 (\textit{resp.} 510) integration points for the reduced-integration domain of ROM $0$ (\textit{resp.} $1$). Building one local ROM takes approximately 2 hours and 30 minutes.

\subsection{Automatic model recommendation}

In this section, a classifier is trained for the automatic model recommendation task. The 120 temperature fields coming from the Sobol' dataset are used as training data for the classifier. Their labels are determined by finding their closest medoid in terms of the ROM-oriented dissimilarity measure. Hence, for each temperature field of the Sobol' dataset, two dissimilarities are computed: one with the medoid of the first cluster, and one with the medoid of the second cluster. Once trained, the classifier can be evaluated on the 80 labelled temperature fields of the MaxProj dataset.

\subsubsection{Feature selection}

\begin{figure}[!h]
\centering
\includegraphics[scale=0.4]{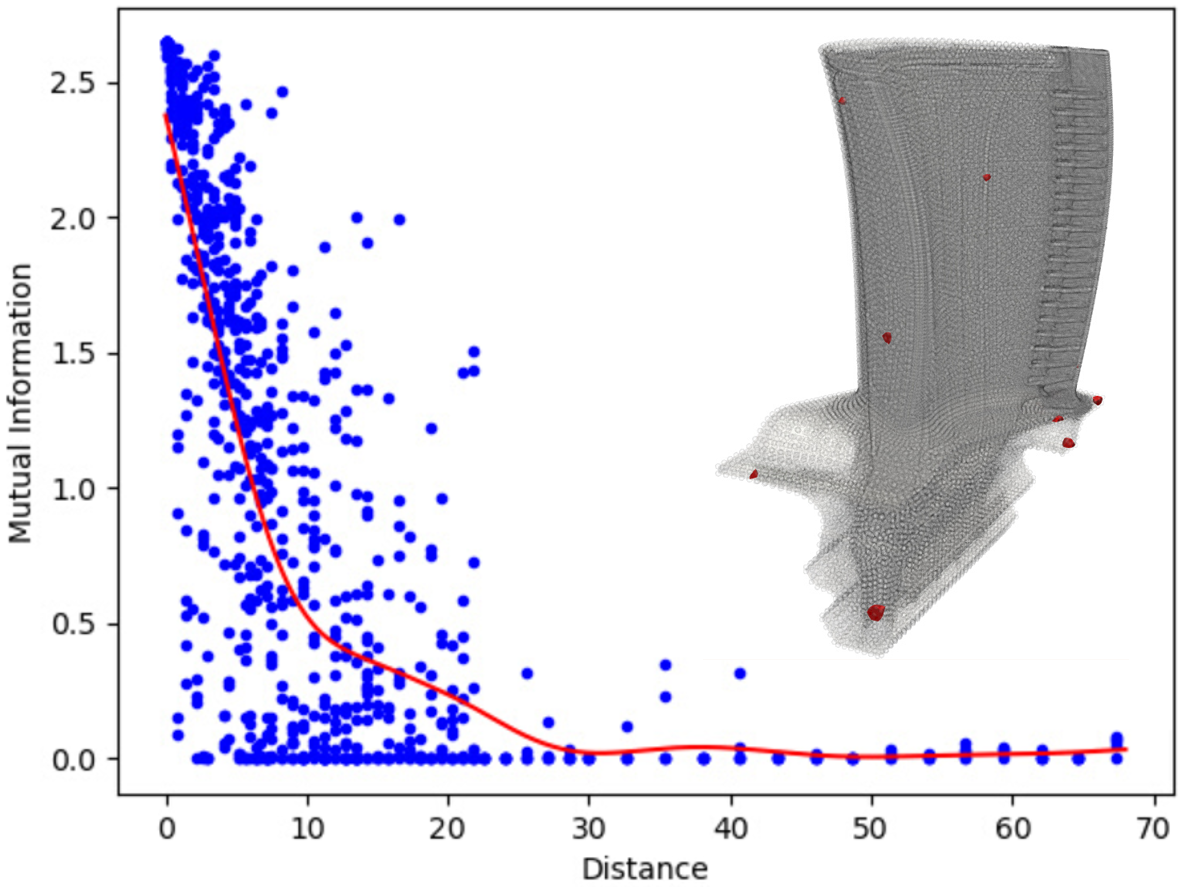}
\caption{Feature selection results. The kriging metamodel for redundancy terms is represented by the red curve and built from 800 true redundancy terms (blue points). The elements containing the selected nodes are represented in the turbine blade geometry.}
\label{FeatureSelectionRecap}
\end{figure}

Each temperature field is discretized on the finite-element mesh, which contains in the order of the million nodes. To reduce the dimension of the input space and facilitate the training phase of the classifier, we apply the geostatistical mRMR feature selection algorithm described in~\cite[Algorithm~1]{mca26010017} on data from the Sobol' dataset. First, 800 pairs of nodes are selected in the mesh, which takes 18 seconds. The 800 corresponding redundancy terms are computed with Scikit-Learn~\cite{scikit-learn} in less than 3 seconds. Figure~\ref{FeatureSelectionRecap} plots the values of these redundancy terms versus the Euclidean distance between the nodes. We observe that the correlation between the redundancy mutual information terms and the distance between the nodes is poor, with a lot of noise. This can be due to the fact that the random temperature fields have been built using Gaussian random fields on the outer surface with an isotropic correlation function depending on the geodesic distance along the surface rather than the Euclidean distance. Since the turbine blade is a relatively thin structure, two nodes, one on the intrados and another one on the extrados, can be close to each other in the Euclidean distance, but with totally uncorrelated temperature fluctuations because of the large geodesic distance separating them. On the contrary, two points on the same side of the turbine blade can have correlated temperature variations while being separated by a Euclidean distance in the order of the blade's thickness. The length of the mutual information's high-variance regime seems to correspond to the blade's chord, which supports this explanation. The thinness of the turbine blade induces anisotropy in the correlation function of the bulk Gaussian random field defining the thermal loading, which implies an anisotropic behavior of the mutual information according to~\cite[Property~1]{mca26010017}. The use of a local temperature perturbation $\delta T_0$ in conjunction with fluctuation modes having larger length scales may also partially explain the large variance of redundancy terms. Nonetheless, it remains clear that redundancy terms are smaller as for large distances. This trend is captured by a kriging metamodel (Gaussian process regression) trained with Scikit-Learn in a few seconds, with a sum-kernel involving the Mat\'{e}rn kernel with parameter $5/2$ (to get a continuous and twice differentiable metamodel) and length scale $1$, and a white kernel to estimate the noise level of the signal. The curve of the metamodel is given in Figure~\ref{FeatureSelectionRecap}. Then, for each node of the finite-element mesh, the mutual information with the label variable is computed. The computations of these relevance terms (in the order of the million terms) are distributed between 280 cores, which gives a total computation time of 15 minutes. Among these features, $5,986$ features are preselected by discarding those with a relevance mutual information lower than $0.05$. The geostatistical mRMR selects $11$ features in 42 seconds. The corresponding nodes in the finite-element mesh can be visualized in Figure~\ref{FeatureSelectionRecap}.

\begin{remark}
The metamodel for redundancy terms could be improved by defining it as a function of the precomputed geodesic distances along the outer surface rather than the Euclidean distances. Each finite-element node would be associated to its nearest neighbor on the outer surface before computing the approximate mutual information from geodesic distances.
\end{remark}

\subsubsection{Classification}

The classifier is trained on the Sobol' dataset, using the values of the temperature fields at the $11$ nodes identified by the feature selection algorithm. The classifier is a logistic regression~\cite{10.2307/2280041, 10.2307/2983890, multinomialLR} with elastic net regularization~\cite{10.2307/3647580} implemented in Scikit-Learn. The two hyperparameters involved in the elastic net regularization are calibrated using 5-fold cross-validation, giving a value of $0.001$ for the inverse of the regularization strength, and $0.4$ for the weight of the $L^1$ penalty term (and thus $0.6$ for the $L^2$ penalty term). Thanks to the $L^1$ penalty term, the classifier only uses $5$ features among the $11$ input features. The classifier's accuracy, evaluated on the MaxProj dataset to use new unseen data, reaches $98.75 \%$. The confusion matrix indicates that $100 \%$ of the test examples belonging to class $0$ have been correctly labeled, and that $2.38\%$ of the test examples belonging to class $1$ have been misclassified. Table~\ref{ClassificationResultsBlade} summarizes the values of precision, recall and F1-score on test data.

\begin{table}[h!]
\begin{center}
\caption{Classification results.}
      \begin{tabular}{ccccc}
        \hline
        Class          &Precision&Recall&F1-score&Support\\ \hline
        0              & $0.9744$  &$1.0000$&$0.9870$  &$38$  \\
        1              & $1.0000$  &$0.9762$&$0.9880$  &$42$  \\  \hline
        Accuracy      & -       &-     &$0.9875$  &$80$  \\
        Macro avg      & $0.9872$ &$0.9881$&$0.9875$  &$80$  \\
        Weighted avg   & $0.9878$ &$0.9875$&$0.9875$  &$80$  \\ \hline
      \end{tabular}
      \label{ClassificationResultsBlade}
\end{center}
\end{table}

\subsection{Surrogate model for Gappy reconstruction}
\label{sec:GappyMetaModel}

When using hyper-reduction, the ROM calls the constitutive equations solver only at the integration points belonging to the reduced-integration domain. It is recalled that the ECM selected 506 (\textit{resp.} 510) integration points for the reduced-integration domain of ROM $0$ (\textit{resp.} $1$), and that the finite-element mesh initially contains a number of integration points in the order of the million. Therefore, after a reduced simulation, dual variables defined at integration points are known only at integration points of the reduced-integration domain. To retrieve the full field, the Gappy-POD~\cite{GappyPOD} finds the coefficients in the POD basis that minimize the squared error between the reconstructed field and the ROM predictions on the reduced-integration domain. This minimization problem defines the POD coefficients as a linear function of the predicted values on the reduced-integration domain. Although these coefficients are optimal in the least squares sense, they can be biased by the errors made by the ROM. To alleviate this problem, we propose to replace the common Gappy-POD procedure by a metamodel or \textit{Gappy surrogate}. The inputs and the outputs of the Gappy surrogate are the same as for the Gappy-POD: the input is a vector containing the values of a dual variable on the reduced-integration domain, and the output is a vector containing the optimal coefficients in the POD basis. One Gappy surrogate must be built for each dual variable of interest: in our case, 7 surrogate models per cluster are required, namely one for the quantity of interest $p_{\textrm{cum}}^{o}$ and one for every component of the Cauchy stress tensor.

The training data for these Gappy surrogates are obtained by running reduced simulations with the local ROMs, using the thermal loadings of the Sobol' dataset. Indeed, the two local ROMs have been built on the MaxProj dataset, therefore thermal loadings of the Sobol' dataset can play the role of test data for the ROMs. For each thermal loading in the Sobol' dataset, the true high-fidelity solution is already known since it has been computed to provide training data for the classifier. In addition, the exact labels for these thermal loadings are known, which means that we know which local ROM to choose for each thermal loading of the Sobol' dataset. Given ROM predictions on the reduced-integration domain, the optimal coefficients in the POD basis are given by the projections of the true prediction made by the high-fidelity model (the finite-element model) onto the POD modes. This provides the true outputs for the Sobol' dataset, which can then be used as a training set for the Gappy surrogates.

Given the high-dimensionality of the input data (there are more than 500 integration points in the reduced-integration domains) with respect to the number of training examples (120 examples), a multi-task Lasso metamodel is used. The hyperparameter controlling the regularization strength is optimized by 5-fold cross-validation. Training the 14 Gappy surrogates (7 for each cluster) takes 1 hour. The Gappy surrogates select between $8\%$ and $18\%$ of the integration points in the reduced-integration domains, thanks to the $L^1$ regularization. The mean cross-validated coefficients of determination are $0.9637$ (\textit{resp.} $0.8935$) for the quantity of interest for cluster $0$ (\textit{resp.} cluster $1$), and range from $0.9404$ to $0.9938$ for stress components. These satisfying results mean that it is not required to train a kriging metamodel with the variables selected by Lasso to get nonlinear Gappy surrogates. The Gappy surrogates are then linear, just as the Gappy-POD.

\begin{remark}
In this strategy, the local ROMs solve the equations of the mechanical problem, which enables using linear surrogate models to reconstruct dual variables. Using surrogate models from scratch instead of local ROMs would have been more difficult, given the nonlinearities of this mechanical problem and the lack of training data for regression. In addition, such surrogate models would require a parametrization of the input temperature fields, whereas the local ROMs use the exact values of the temperature fields on the RID without assuming any model for the thermal loading.
\end{remark}

The dictionary-based ROM-net used for mechanical simulations of the high-pressure turbine blade is made of a dictionary of two local hyper-reduced order models and a logistic regression classifier. The classifier analyzes the values of the input temperature field at 11 nodes only, identified by our feature selection strategy. For a given thermal loading in the exploitation phase, after the reduced simulation with the local ROM recommended by the classifier, linear cluster-specific Gappy surrogates reconstruct the full dual fields (quantity of interest and stress components) from their predicted values on the reduced-integration domain.

\subsection{Uncertainty quantification results}
\label{sec:UQres}


Once trained, the ROM-net can be applied for the quantification of uncertainties on the mechanical behavior of the HP turbine blade resulting from the uncertainties on the thermal loading. Since the ROM-net online operations can be performed sequentially on one single core, 24 cores are used in order to compute the solution for 24 thermal loadings at once. This way, 42 batches of 24 Monte Carlo simulations are run in 2 hours and 48 minutes, using Safran's module of \textit{Mordicus} code. The 1008 thermal loadings used for this study are generated by randomly sampling points from the uniform distribution on the 5D unit hypercube and applying the transformation given in Equation~\eqref{TransfoUniformToTempReducedCoords}.

\begin{figure}[!h]
\centering
\includegraphics[scale=0.35]{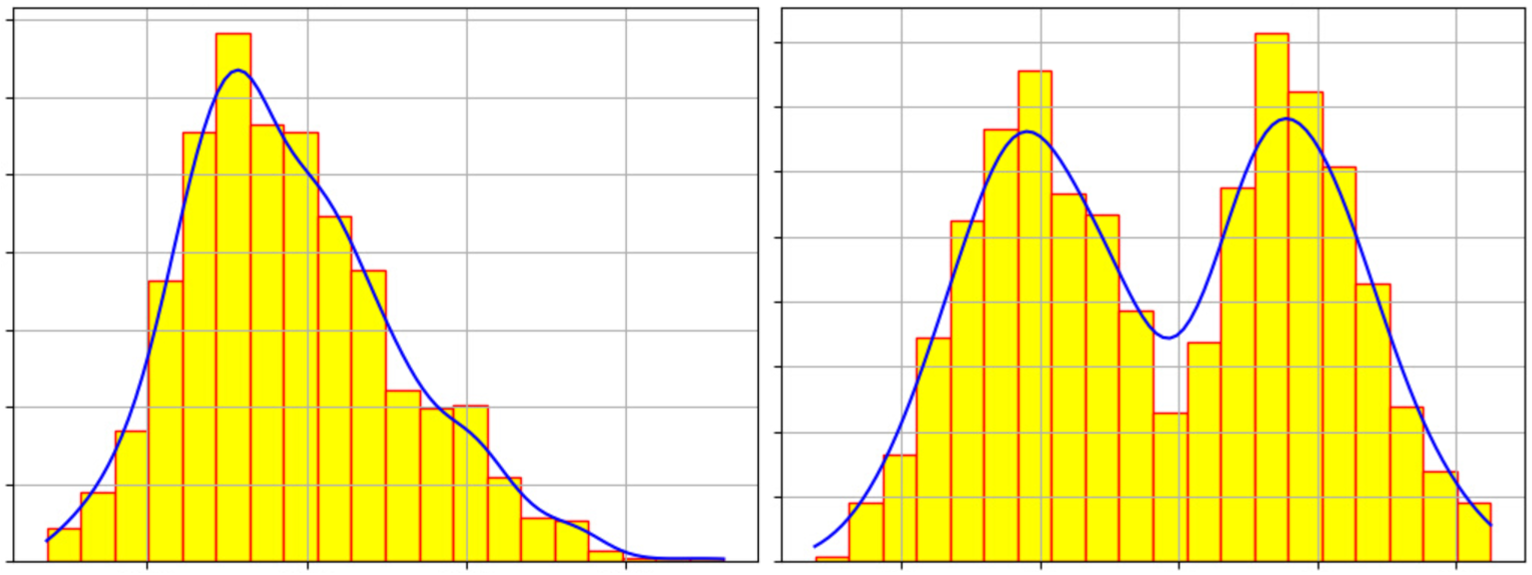}
\caption{Histograms and probability density functions of the quantities of interest $\overline{p}_{\textrm{cum}}^o$ (left) and $\overline{\sigma}_{\textrm{eq}}$ (right).}
\label{UQresults}
\end{figure}

\begin{table}[h!]
\begin{center}
\caption{Widths of the confidence intervals (CI) for the expectations, expressed as percentages of the estimated expectations.}
      \begin{tabular}{ccc}
        \hline
        Estimated variable & Confidence level & Relative CI width \\ \hline
        $\E[\overline{p}_{\textrm{cum}}^o]$ & $0.95$  & $2.16 \%$  \\
        $\E[\overline{p}_{\textrm{cum}}^o]$ & $0.99$  & $2.84 \%$  \\
        $\E[\overline{\sigma}_{\textrm{eq}}]$ & $0.95$  & $1.26 \%$  \\
        $\E[\overline{\sigma}_{\textrm{eq}}]$ & $0.99$  & $1.66 \%$  \\ \hline
      \end{tabular}
      \label{TableCIwidths}
\end{center}
\end{table}

The expected values of $\overline{p}_{\textrm{cum}}^o$ and $\overline{\sigma}_{\textrm{eq}}$ are estimated with the empirical means $\overline{Z}_n=\frac{1}{n}\sum_{i=1}^n Z_i$, where $Z_i$ are the corresponding samples. The variances of $\overline{p}_{\textrm{cum}}^o$ and $\overline{\sigma}_{\textrm{eq}}$ are computed using the unbiased sample variance $\displaystyle S_n^2=\frac{1}{n-1}\sum_{i=1}^n\left(Z_i-\overline{Z}_n\right)^2$. The Central Limit Theorem gives asymptotic confidence intervals for the expected values: for all $\alpha \in ]0;1[$, the interval:\begin{equation}
I_n = \left[ \overline{Z}_n - \phi_{1-\frac{\alpha}{2}}\sqrt{S_{n}^{2}/n}; \overline{Z}_n + \phi_{1-\frac{\alpha}{2}} \sqrt{S_{n}^{2}/n},   \right]
\label{ConfidenceIntervalDef}
\end{equation}
where $\phi_r$ denotes the quantile of order $r$ of the standard normal distribution $\mathcal{N}(0,1)$ is an asymptotic confidence interval with confidence level $1-\alpha$ for the expectation $\mu$: $\lim_{n \rightarrow +\infty} \Prob(\mu \in I_n ) = 1 - \alpha$.
The widths of the confidence intervals are expressed as a percentage of the estimated value for the expectations in Table~\ref{TableCIwidths}.

The probability density functions of the quantities of interest can be estimated using Gaussian kernel density estimation (see Section~6.6.1. of~\cite{Hastie2005TheEO}). Figure~\ref{UQresults} gives the histograms and estimated distributions for $\overline{p}_{\textrm{cum}}^o$ and $\overline{\sigma}_{\textrm{eq}}$. The shapes of these distributions highly depend on the assumptions made for the stochastic thermal loading. As observed in Figure~\ref{QoIRef}, the stress field is highly sensitive to temperature gradients, which may explain why the distribution of the Von Mises stress is bimodal.

\subsection{Workflow}
\label{sec:worfklow}

\begin{figure}[!h]
\centering
\includegraphics[scale=0.45]{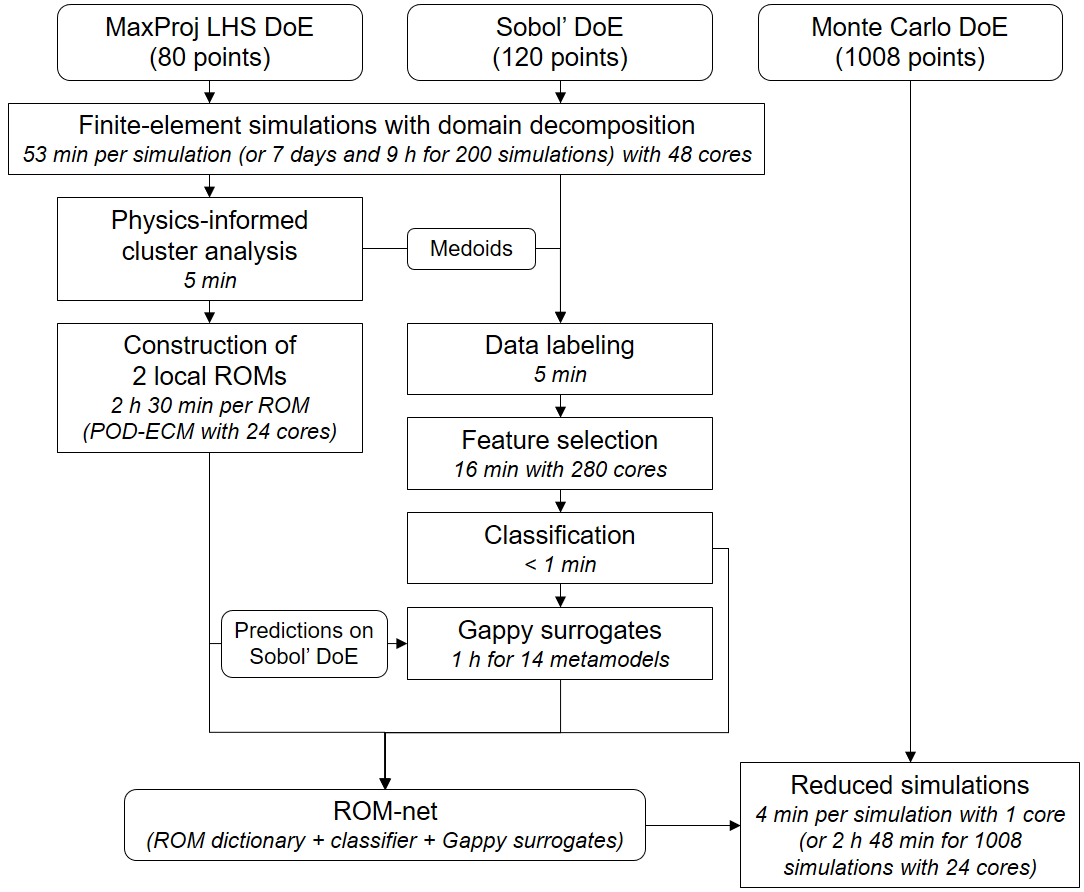}
\caption{Workflow for the ROM-net methology applied to the considered industrial setting in Sections~\ref{sec:does}-\ref{sec:UQres}.}
\label{workflow}
\end{figure}

The workflow presented in Sections~\ref{sec:does}-\ref{sec:UQres} is illustrated in Figure~\ref{workflow}.

\subsection{Validation}

\begin{table}[!h]
\begin{center}
\caption{Error indicators for the evaluation of the ROM-net on 20 new thermal loadings.}
      \begin{tabular}{ccc}
        \hline
        Error indicator & Errors on $p_{\textrm{cum}}^o$ & Errors on $\sigma_{\textrm{eq}}$ \\ \hline
        Mean $L^2$ relative error on $\Omega$ & $1.14 \%$  & $0.84 \%$  \\
        Mean $L^2$ relative error on $\Omega'$ & $0.75 \%$  & $1.46 \%$  \\
        Mean $L^{\infty}$ relative error on $\Omega$ & $1.11 \%$  & $1.09 \%$  \\
        Mean $L^{\infty}$ relative error on $\Omega'$ & $1.05 \%$  & $2.60 \%$  \\ 
        Mean relative error on value averaged over $\Omega'$ & $0.50 \%$  & $0.89 \%$  \\
        Mean distance between maxima & $0$  & $0$  \\ \hline
      \end{tabular}
      \label{ROM-netErrors}
\end{center}
\end{table}

For validation purposes, the accuracy of the ROM-net is evaluated on 20 Monte Carlo simulations with 20 new thermal loadings. These thermal loadings are generated by randomly sampling points from the uniform distribution on the 5D unit hypercube, and applying the transformation given in Equation~\eqref{TransfoUniformToTempReducedCoords}. The reduced simulations are run on single cores with Safran's module of \textit{Mordicus} code. The total computation time for generating a new thermal loading on the fly, selecting the corresponding reduced model, running one reduced simulation and reconstructing the quantities of interest is 4 minutes on average. As a comparison, one single high-fidelity simulation with \textit{Z-set}~\cite{zset} with 48 subdomains takes 53 minutes, which implies that the ROM-net computes $13.25$ times faster. However, one high-fidelity simulation requires 48 cores for domain decomposition, whereas the ROM-net works on one single core. Hence, using 48 cores to run 48 reduced simulations in parallel, $636$ reduced simulations can be computed in 53 minutes with the ROM-net, while the high-fidelity model only runs one simulation. In addition to the acceleration of numerical simulations, energy consumption is reduced by a factor of $636$ in the exploitation phase. In spite of the fast development of high-performance computing, numerical methods computing approximate solutions at reduced computational resources and time are particularly important for many-query problems such as uncertainty quantification, where the intensive use of computational resources is a major concern. Model order reduction and ROM-nets play a prominent role toward \textit{green} numerical simulations~\cite{doi:10.3166/ejcm.19.365-388}. Of course, the number of simulations in the exploitation phase must be large enough to compensate the efforts made in the training phase, like in any machine learning or model order reduction problem.

\begin{figure}[!h]
\centering
\includegraphics[scale=0.45]{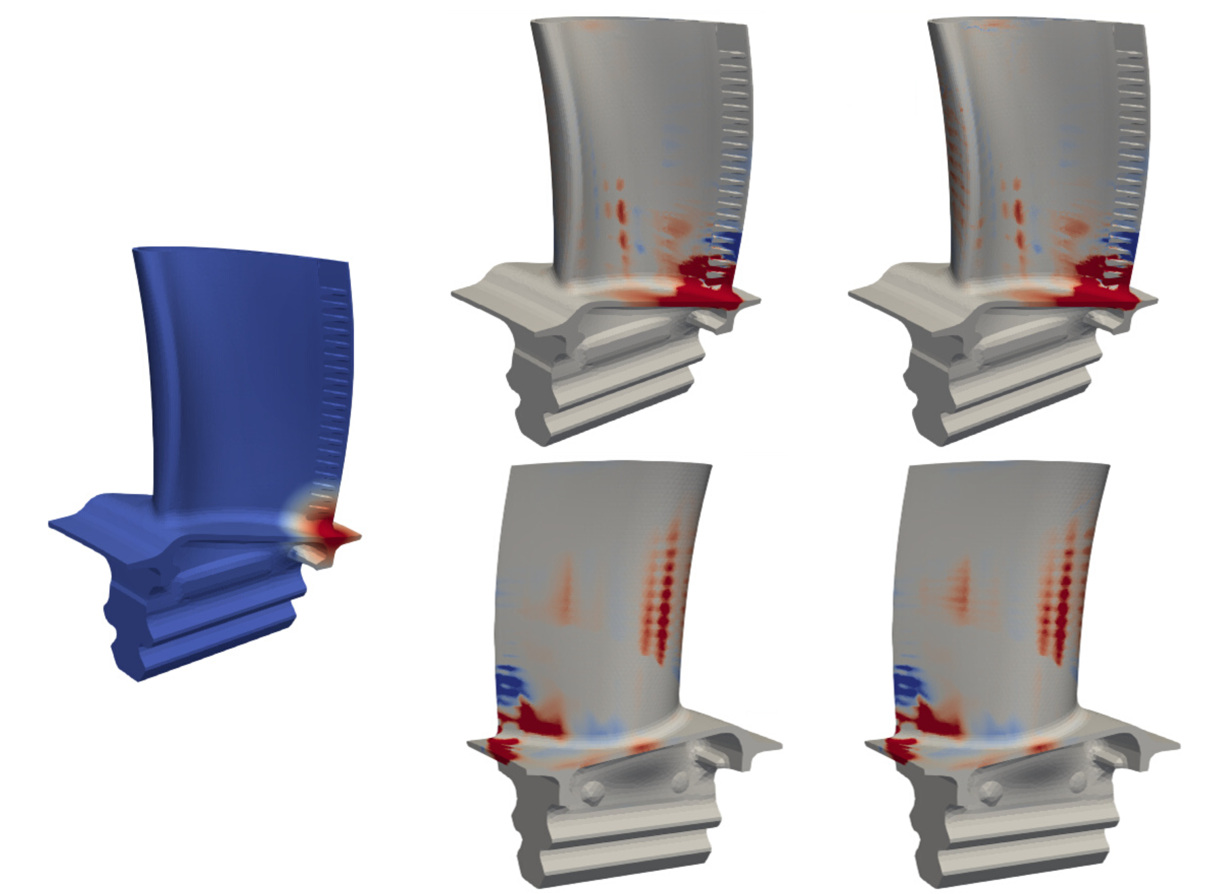}
\caption{Comparison between high-fidelity predictions (middle column) and ROM-net's predictions (right-hand column). The field on the left represents the difference between the current temperature field (belonging to cluster $0$) and the reference one. The other fields correspond to the increments of the quantity of interest $p_{\textrm{cum}}^{o}$ with respect to its reference state obtained with the reference temperature field.}
\label{Comparison0ROM-net}
\end{figure}

\begin{figure}[!h]
\centering
\includegraphics[scale=0.45]{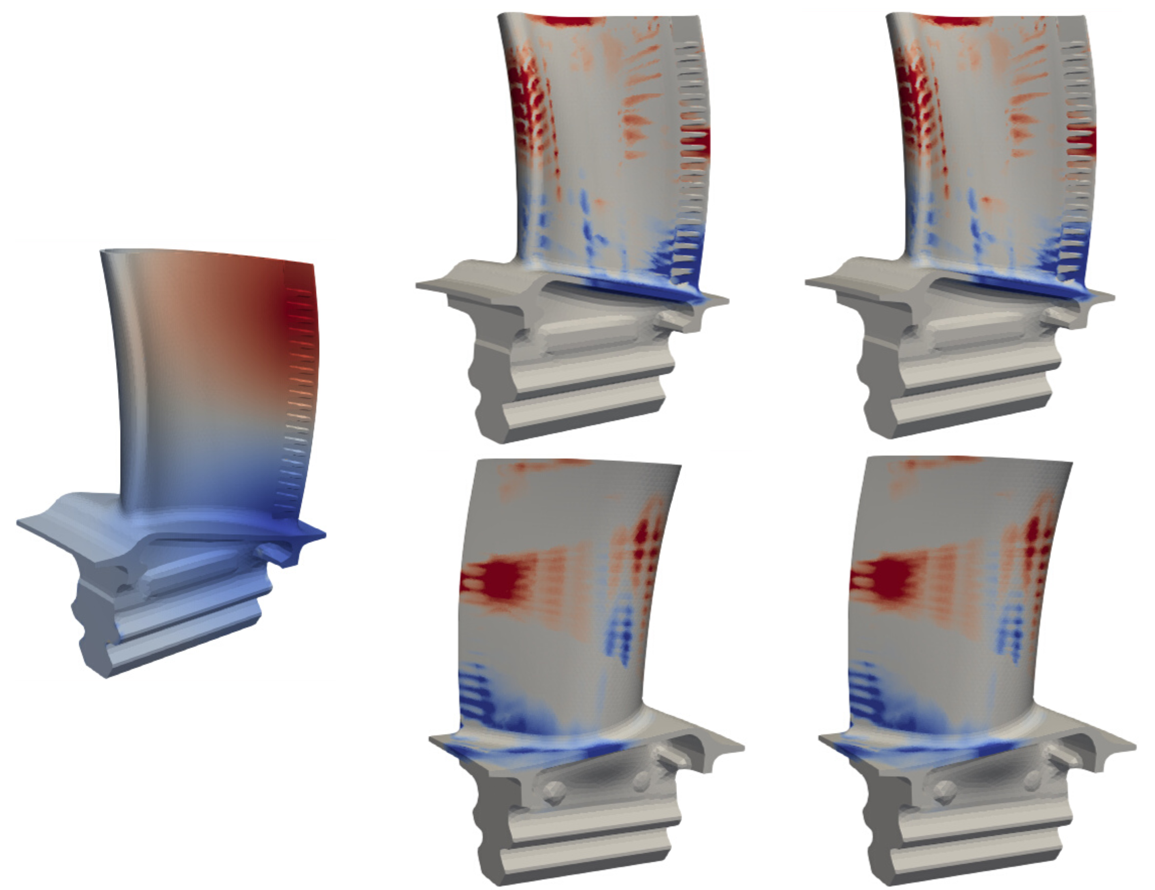}
\caption{Comparison between high-fidelity predictions (middle column) and ROM-net's predictions (right-hand column). The field on the left represents the difference between the current temperature field (belonging to cluster $1$) and the reference one. The other fields correspond to the increments of the quantity of interest $p_{\textrm{cum}}^{o}$ with respect to its reference state obtained with the reference temperature field.}
\label{Comparison1ROM-net}
\end{figure}

Figures~\ref{Comparison0ROM-net} and~\ref{Comparison1ROM-net} show the results for two simulations belonging to cluster $0$ and cluster $1$ respectively. These figures give the difference between the current temperature field as the reference one, \textit{i.e.} the field $T-T_{\textrm{ref}}$, and the resulting variations of the quantity of interest predicted by the ROM-net and the high-fidelity model, \textit{i.e.} $p_{\textrm{cum}}^{o,\textrm{ROM}}(T) - p_{\textrm{cum}}^{o,\textrm{HF}}(T_{\textrm{ref}})$ and $p_{\textrm{cum}}^{o,\textrm{HF}}(T) - p_{\textrm{cum}}^{o,\textrm{HF}}(T_{\textrm{ref}})$. The signs and the positions of the variations of the quantity of interest seem to be quite well predicted by the ROM-net.

\begin{figure}[!h]
\centering
\includegraphics[scale=0.3]{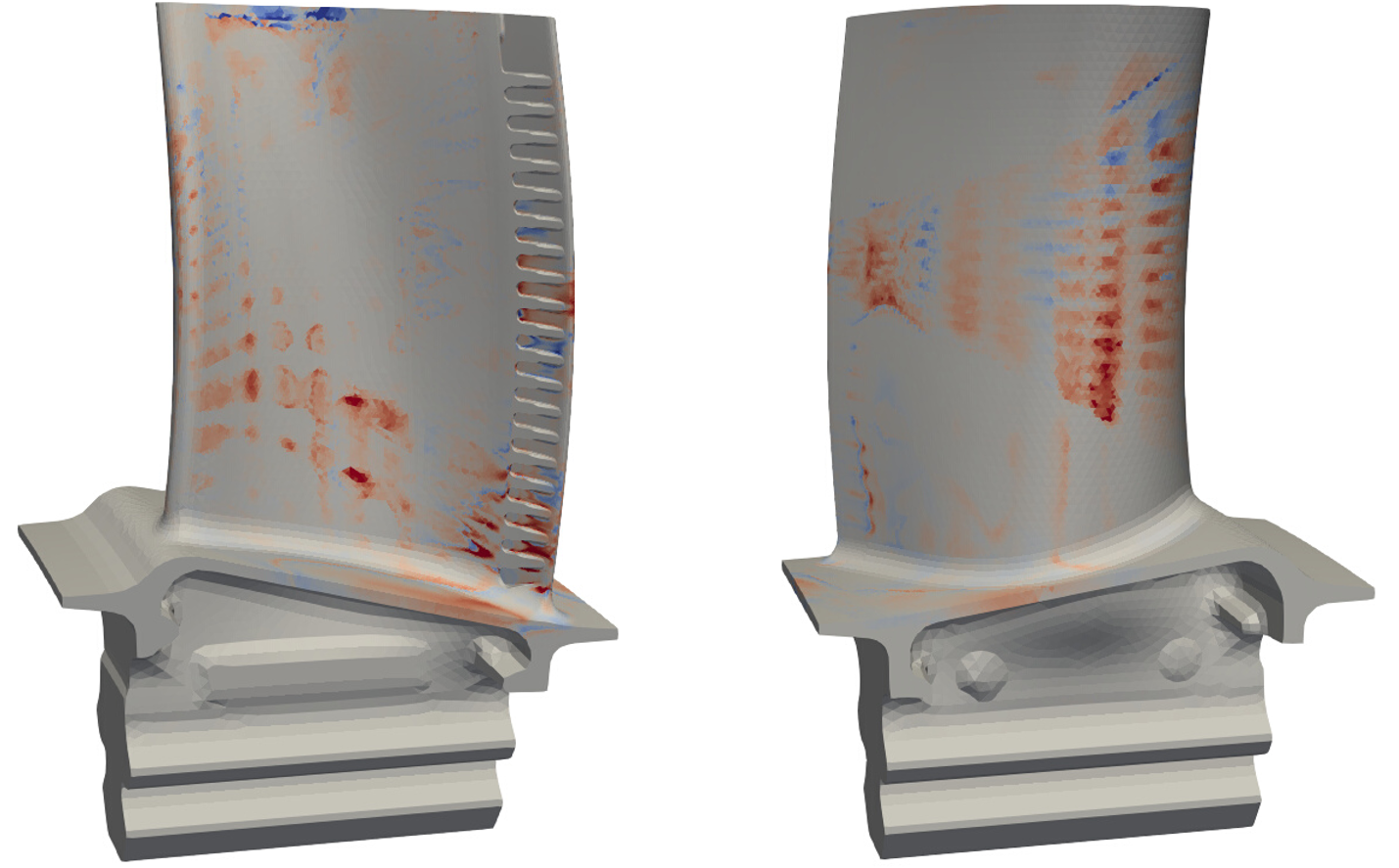}
\caption{Errors on the quantity of interest $p_{\textrm{cum}}^{o}$. The red (\textit{resp.} blue) color is used for zones where the quantity of interest is overestimated (\textit{resp.} underestimated).}
\label{ErrorField}
\end{figure}

Let us introduce a zone of interest $\Omega'$ defined by all of the integration points at which $p_{\textrm{cum}}^o$ is higher than $0.4 \times \max p_{\textrm{cum}}^{o} (\bs{\xi})$ for the thermal loading defined by $T_{\textrm{ref}} + \delta T_0$. This zone of interest contains 209 integration points. The values of the variables $p_{\textrm{cum}}^o$ and $\sigma_{\textrm{eq}}$ averaged over $\Omega'$ are denoted by $\overline{p}_{\textrm{cum}}^o$ and $\overline{\sigma}_{\textrm{eq}}$. Table~\ref{ROM-netErrors} gives different indicators quantifying the errors made by the ROM-net: the $L^2$ relative errors on the whole domain $\Omega$ and on the zone of interest $\Omega'$, the $L^{\infty}$ relative errors on $\Omega$ and $\Omega'$, the relative errors on $\overline{p}_{\textrm{cum}}^o$ and $\overline{\sigma}_{\textrm{eq}}$, and the errors on the locations of the points where the fields $p_{\textrm{cum}}^o$ and $\sigma_{\textrm{eq}}$ reach their maxima. All the relative errors remain in the order of $1 \%$ or $2 \%$, which validates the methodology. In addition, the ROM-net perfectly predicts the position of the critical points at which $p_{\textrm{cum}}^o$ and $\sigma_{\textrm{eq}}$ reach their maxima. Figure~\ref{ErrorField} shows errors on the quantities of interest.

\section{Conclusion}
\label{sec:conclusion}

In this work, we used a dictionary-based ROM-net to successfully quantify the uncertainties on dual quantities of interest of an elastoviscoplastic high-pressure turbine blade, generated by the uncertainty on its temperature loading. 
This validates the methodology on large models with highly nonlinear behaviors. An outlook of this work would be to consider nonparametrized geometrical variability, which is of paramount interest when considering the design of mechanical parts and the uncertainty quantification of their manufacturing processes.

\vspace{0.3cm}

\noindent \textbf{Acknowledgment:~}{The authors wish to thank Sébastien Da Veiga and Clément Bénard for fruitful discussions on uncertainty quantification.}

\vspace{0.3cm}

\noindent \textbf{Funding:~}{This research was partially funded by the French Fonds Unique Interministériel (MOR\_DICUS).}

\bibliographystyle{plain}
\bibliography{biblio}

\end{document}